\documentclass{article}

\usepackage{arxiv}

\usepackage[utf8]{inputenc} 
\usepackage[T1]{fontenc}    
\usepackage{hyperref}       
\usepackage{url}            
\usepackage{booktabs}       
\usepackage{amsfonts}       
\usepackage{nicefrac}       
\usepackage{microtype}      
\usepackage{lipsum}

\usepackage{times}
\usepackage{epsfig}
\usepackage{graphicx}
\usepackage{amsmath}
\usepackage{amssymb}

\usepackage{adjustbox}

\usepackage{colortbl}
\usepackage{booktabs}
\usepackage[table,x11names,dvipsnames,xcdraw]{xcolor}

\title{Planar geometry and image recovery from motion-blur}

\author{Kuldeep Purohit$^{1}$ \qquad Subeesh Vasu$^{2}\thanks{Work done while at Indian Institute of Technology Madras, India.}$ \qquad M. Purnachandra Rao$^{1}$ \qquad A. N. Rajagopalan$^{1}$ \\
$^1$ Indian Institute of Technology Madras, India \hspace{1cm}
$^2$ \'Ecole polytechnique f\'ed\'erale de Lausanne~(EPFL)\\
{\tt\small kuldeeppurohit3@gmail.com, subeeshvasu@gmail.com, mpurna2u@gmail.com, raju@ee.iitm.ac.in}
} 

\date{}
\begin{document}
\maketitle

\begin{abstract}
Existing works on motion deblurring either ignore the effects of depth-dependent blur or work with the assumption of a multi-layered scene wherein each layer is modeled in the form of fronto-parallel plane. In this work, we consider the case of 3D scenes with piecewise planar structure i.e., a scene that can be modeled as a combination of multiple planes with arbitrary orientations. We first propose an approach for estimation of normal of a planar scene from a single motion blurred observation. We then develop an algorithm for automatic recovery of number of planes, the parameters corresponding to each plane, and camera motion from a single motion blurred image of a multiplanar 3D scene. Finally, we propose a first-of-its-kind approach to recover the planar geometry and latent image of the scene by adopting an alternating minimization framework built on our findings. Experiments on synthetic and real data reveal that our proposed method achieves state-of-the-art results.
\end{abstract}


\section{Introduction}
{R}{ecovery} of 3D structure from images is an extensively researched area in computer vision. Algorithms for scene geometry recovery find applications in visual servoing, video conferencing, tracking, active vision, augmented reality etc. Well-known cues for depth recovery include disparity \cite{lee2011real}, optical flow \cite{shahraray1988robust}, texture \cite{super1995planar, brown1990surface}, shading \cite{zhang1999shape}, defocus blur \cite{chaudhuri2012depth}, and motion blur \cite{chandramouli2010inferring, lin2006depth, zheng2011structure}. While depth estimation has been of general interest, some of the works in literature target the case of inferring piecewise planar geometry (Manhattan model). This was primarily motivated by the fact that the world around us can, in many cases, be modeled as being piecewise planar. Estimating a 3D geometry in terms of planar parameters has tremendous advantages including reduction in the computational complexity and robustness to pixel-level errors in depth cues.

Many works exist in the literature that specifically addresses the task of inferring planar scene geometry from a single image. To recover the surface orientation, foreshortening of texture was used as a cue in \cite{brown1990surface} whereas \cite{super1995planar} used local variations of spatial frequencies. The orientation of text planes was estimated using perspective geometry in \cite{clark2001estimating}. The work in \cite{farid2007estimating} revealed the fact that higher-order correlations in the frequency domain caused by the projection of a planar texture are proportional to the orientation of the plane. \cite{greiner2010estimation} proposed a method to determine the surface normal using projective geometry and spectral analysis. While all the above methods work under the assumption of clean images, there exist very few works which attempt to make use of the cues from degradations (in the form of blur) to estimate plane normal. In \cite{mccloskey2009planar}, optical blur is used as a cue to estimate the planar orientation from a single image. The works in \cite{rao2014inferring}, \cite{vasu2015tapping} utilize \emph{motion blur} to infer the surface normal of the scene from a single motion blurred image, but by assuming the case of in-plane translational camera motion.

There have been few attempts to estimate the complete 3D structure of scene from a single image using learning-based approaches. The work in \cite{saxena2009make3d} used a Markov Random Field trained via supervised learning to infer a set of plane parameters associated with the scene. \cite{haines2012detecting} proposed an approach to identify multiple distinct planes, and estimating their orientation from a single image of an outdoor urban scene by learning the relationship between appearance and structure from a large set of labeled examples. 
 
Lately, convolutional neural networks (CNN) are being increasingly used to address the ill-posedness of single image depth estimation. They are trained on specific datasets formed with the help of multi-view images or depth sensors \cite{eigen2014depth,laina2016deeper,li2018megadepth} to predict depth map from a single image. However, the performance of these methods degrades on general test images that are different from the labeled data available during training. Moreover, accurate depth estimation becomes a challenge in the presence of blur since the fine-level depth cues get subdued in the presence of motion blur.

Motion blurred images have attracted increased attention in research \cite{rao2014harnessing, mohan2019unconstrained, nimisha2018generating, vasu2017local, paramanand2014shape, paramanand2011depth, vijay2013non, nimisha2018unsupervised, purohit2020region, vasu2018non, purohit2019bringing,rajagopalan2005background}, owing to the ubiquity of mobile phones and hand-held imaging devices. Recent years have witnessed significant progress in single image motion deblurring. While the standard blind deblurring algorithms such as \cite{fergus2006removing,cho2009fast,xu2010two,sun2013edge,michaeli2014blind} consider the motion blur to be uniform across the image, various methods have been proposed to handle blur variations due to camera rotational motion \cite{gupta2010single,hirsch2011fast,whyte2012non,Vasu_2017_CVPR,Yan_2017_CVPR} and scene depth variations \cite{hu2014joint,seemakurthy2016deskewing}. However, none of the existing approaches address multi-planar inclined scenes.

Although the problem of blur and depth estimation are individually quite challenging, a few attempts have been made in the literature to jointly tackle the two problems. Among the existing methods on motion deblurring, the ones that come close to that of ours is \cite{hu2014joint} and \cite{seemakurthy2016deskewing}. The work in \cite{hu2014joint} have proposed to jointly estimate depth and non-uniform blur from a single blurred image but is designed to handle only piecewise fronto-parallel planar scenes. \cite{seemakurthy2016deskewing} was designed to remove the motion blur effects caused in underwater imaging by modeling it via a virtual depth map characterized using a single exponential function. While few other works on depth-aware motion deblurring such as \cite{sorel2008space,xu2012depth,paramanand2013non} have also been proposed, they rely on multiple observations.

Recently, various learning based attempts have been proposed to solve the problem of removing heterogeneous blur from a single blurred image. \cite{sun2015learning} trained a CNN for predicting a probability distribution of motion blur at the patch level. To recover the latent image, \cite{gong2017motion} estimated a dense motion flow with a fully convolutional neural network. \cite{nimisha2017blur} used adversarial training to learn blur-invariant features to perform motion deblurring. End-to-end trainable multi-scale CNN models are proposed in \cite{nah2017deep,tao2018scale} to restore the latent image directly. Although the above methods attempt to solve the deblurring problem in more generic settings, the performance of these methods depends purely on the training data and the learning capability of the underlying networks. While the learning based models have been shown to handle a few types of heterogeneous blur, their performance on generic blurred images is not guaranteed. At the same time, the performance of some of these methods on standard datasets such as \cite{kohler2012recording} reveal the fact that conventional methods still outperform the learning based approaches when it comes to specific image formation models.

In this paper, we not only extend our previous work in \cite{rao2014inferring} and but also bring in many other contributions. First, we show how the approach in \cite{rao2014inferring} can be modified to account for the camera motions involving rotations too. We then develop a fully-automatic first-of-its-kind approach to recover the number of planes, the parameters corresponding to each plane, and the camera motion from a single motion blurred image of a scene with multiple planes. These results are then used to pose an alternating minimization problem to recover the complete scene geometry as well as the latent image of the scene. On motion blurred images, our depth estimates are more accurate than learning based approaches \cite{laina2016deeper,li2018megadepth} due to the cues present in the blur-kernels and additional constraints present in the algorithm of motion deblurring. In this paper, we relax majority of constraints that were being enforced in previous works. Unlike \cite{rao2014inferring} which handles the case of in-plane translational motion of the camera alone, our proposed approach for normal estimation can handle more general kinds of camera motion. In addition, we also tackle the case of multi-planar scenes and propose a novel formulation for deblurring of such scenes. Unlike \cite{hu2014joint} which requires user interaction and relies on piecewise fronto-parallel planar assumption, our approach is fully automatic and uses only a piecewise planar representation of the scene.

The key contributions of our work are summarized below
\begin{itemize}
\item This is the first work in literature to perform surface normal estimation from general motion blur present in a single image.
\item We develop a fully-automatic algorithm to estimate the number of planes, parameters corresponding to each plane, and the camera motion from a single motion blurred image.
\item We propose an elegant alternating minimization approach to jointly estimate the scene geometry and latent image from a single motion blurred image. Our proposed approach is able to deliver state-of-the-art results on single image depth-aware deblurring.
\end{itemize}

The remainder of this paper is organized as follows. Our proposed approach for normal estimation directly from the blur kernels is introduced in Section 2. In Section 3, as an application to our findings, we propose a potential use of the estimated normals to perform blind deblurring of a scene containing multiple inclined planes. This is followed by experimental results in Section 4.

\section{Normal Estimation from Blur Kernels}
\label{normal_est}
This section describes our approach, wherein we employ PSFs extracted from various locations in a motion blurred image to estimate the surface normal of the underlying scene. For the case of the fronto-parallel planar scene, all the blur kernel are one and the same, since all of them are at the same depth and the camera motion contains only in-plane translations. However, for the inclined planar scene, the size of the blur kernel varies with the scene depth, indicating that the blur kernels themselves carry the cue about the surface normal of the underlying scene. This is the key observation which motivated us to formulate a technique where one can determine the surface normal using pixel-shift information contained in the PSFs from different locations in a blurred image. 

In general, the homography $\mathbf{H}_p$ is a function of $6$ dimensional (6D) camera motion (3D rotations and 3D translations). However, recent works in \cite{gupta2010single,kohler2012recording} have shown that the effect of camera motion encountered in practice can be well-approximated using in-plane rotations and translations thereby reducing the space of camera motion from 6D to 3D while not compromising on the validity of image formation model. Hence, we too adopt this approximation to reduce the ill-posedness of the associated problems that we are going to address. Thus we use the homography $\mathbf{H}_p$ which is parameterized by translation along $X$-axis ($t_{X_p}$) and $Y$-axis ($t_{Y_p}$), and rotations about $Z$-axis ($\theta_{Z_p}$). Therefore the equation for $\mathbf{R}_p$  can be simplified to the following form
\begin{equation}
\label{rotn_matrix_eqn}
\mathbf{R}_p = \left[ {\begin{array}{ccc}
 cos(\theta_{Z_p}) & sin(\theta_{Z_p}) & 0  \\
 -sin(\theta_{Z_p}) & cos(\theta_{Z_p}) & 0  \\
 0 & 0 & 1  \\
\end{array} } \right]
\end{equation}
Furthermore, a recent work \cite{Vasu_2017_CVPR} has shown that, for typical handshakes the blur induced by in-plane rotations of the camera can be very well modeled with small $\theta _Z$ (i.e; cos ($\theta _Z$) $\cong 1$ and sin($\theta _Z$) $\cong \theta _Z$). Our proposed solution for the normal estimation tries to exploit the linearization capability of small $\theta _Z$ approximation model for rotational motion. Thus for the case of general camera motion, the overall homography matrix can be simplified to the following form.
 \begin{equation}
 \label{small_teta_approx}
 \mathbf{H}_p = \left[ {\begin{array}{ccc}
1+n_X\frac{t_{X_p}}{d} & \theta_{Z_p} + n_Y\frac{t_{X_p}}{d}& \nu n_Z\frac{t_{X_p}}{d}  \\
 -\theta_{Z_p} + n_X\frac{t_{Y_p}}{d} &1+n_Y\frac{t_{Y_p}}{d}  & \nu n_Z\frac{t_{Y_p}}{d} \\
 0 & 0 & 1
\end{array} } \right]
 \end{equation}

For the case of an inclined scene with orientation ${\bf n}=[n_X \ n_Y \ n_Z]^T$, consider a single camera pose $p$ that is involved in the formation of PSF at position $\textbf{x}=(x,y)$. The camera pose $p$ shift the intensity at pixel location $(x,y)$ to a new location $(x_p, y_p)$, which can be determined as
\begin{equation}
\label{inclined_pixel_shift}
\left[ {\begin{array}{c}
x_p \\
y_p \\
1
\end{array} } \right]=  \left[ {\begin{array}{ccc}
1+n_X\frac{t_{X_p}}{d} & \theta_{Z_p} + n_Y\frac{t_{X_p}}{d}& \nu n_Z\frac{t_{X_p}}{d}  \\
 -\theta_{Z_p} + n_X\frac{t_{Y_p}}{d} &1+n_Y\frac{t_{Y_p}}{d}  & \nu n_Z\frac{t_{Y_p}}{d} \\
 0 & 0 & 1
\end{array} } \right]
\left[ {\begin{array}{c}
x \\
y \\
1
\end{array} } \right]
\end{equation}
Eq. \eqref{inclined_pixel_shift} implies that the pixel shifts are no longer a constant, but vary as a function of the spatial coordinates $x$ and $y$. This, in turn leads to variation in the blur kernels as well.

The linearity of the relationship between pixel shifts and the surface normal can be further pronounced, if the quantity being considered is the difference in the shift caused due to two different camera positions $\mathbf{t_{p_{1}}}=[t_{X_{p_{1}}} \ t_{Y_{p_{1}}} \  0]^T$ and $\mathbf{t_{p_{2}}}=[t_{X_{p_{2}}} \ t_{Y_{p_{2}}} \  0]^T$ as in the following relation
\begin{equation}
\label{pixel_shift_nrml_eqn}
\left[ {\begin{array}{c}
\Delta x \\
\Delta y \\
1
\end{array} } \right]=  \left[ {\begin{array}{ccc}
n_X  \dfrac{\Delta t_{X}}{d} & \Delta \theta _Z + n_Y \dfrac{\Delta t_{X}}{d} & \nu n_Z \dfrac{\Delta t_{X}}{d}\\
n_X  \dfrac{\Delta t_{Y}}{d} -\Delta \theta _Z & n_Y \dfrac{\Delta t_{Y}}{d} & \nu n_Z \dfrac{\Delta t_{Y}}{d}\\
 0 & 0 & 1
\end{array} } \right]
\left[ {\begin{array}{c}
x \\
y \\
1
\end{array} } \right]
\end{equation}
where $\Delta x = x_{p_{1}} - x_{p_{2}}$, $\Delta y = y_{p_{1}} - y_{p_{2}}$, $\Delta t_{X} = t_{X_{p_{1}}} - t_{X_{p_{2}}}$, $\Delta t_{Y} = t_{Y_{p_{1}}} - t_{Y_{p_{2}}}$, and $\Delta \theta _Z = \theta_{Z_{p_{1}}}-\theta_{Z_{p_{2}}}$. The relation in Eq. \eqref{pixel_shift_nrml_eqn} can be rearranged to obtain a linear relation between the unknown $\textbf{n}$ and the pixel shifts along $x$ and $y$ direction induced at a location $\textbf{x}$ as
\noindent\begin{minipage}{.5\linewidth}
\begin{equation}
\label{Delta x eqn}
\Delta x
=
\left[ {\begin{array}{r}
x \
y \
1
\end{array} } \right]
 \left[ {\begin{array}{c}
n_X\frac{\Delta t_X}{d} \\ n_Y\frac{\Delta t_X}{d} + \Delta \theta _Z \\ \nu n_Z\frac{\Delta t_X}{d}
\end{array} } \right]
\end{equation}
\end{minipage}%
\begin{minipage}{.5\linewidth}
\begin{equation}
\label{Delta y_eqn}
\Delta y
=
\left[ {\begin{array}{r}
x \
y \
1
\end{array} } \right]
 \left[ {\begin{array}{c}
 n_X\frac{\Delta t_Y}{d} - \Delta \theta _Z \\ n_Y\frac{\Delta t_Y}{d}  \\ \nu n_Z\frac{\Delta t_Y}{d}
\end{array} } \right].
\end{equation}
\end{minipage}

As can be deduced from Eq. \eqref{Delta x eqn} and Eq. \eqref{Delta y_eqn}, unlike the case for pure in-plane translations, the PSFs induced by general camera motion will be spatially varying even for the case of fronto-parallel planar scenes. The blur kernels are no longer spatially invariant for the case of fronto-parallel scene. However, by comparing the kernels from first and second row, it can be observed that the variation induced by the translational camera motion in each blur kernel still carries information about the surface normal.

Although the presence of $\Delta \theta$ preempts the recovery of surface normal directly from the quantities in the right-most column vector of Eq. \eqref{Delta x eqn} alone, we can utilize the information from both Eq. \eqref{Delta x eqn} and Eq. \eqref{Delta y_eqn} together to overcome this issue. Let us denote the entries in the right-most column vector in Eq. \eqref{Delta x eqn} and Eq. \eqref{Delta y_eqn} as $\textbf{b}_{x}=[a_{x} \ b_{x} \  c_{x}]^T$ and $\textbf{b}_{y} = [a_{y} \ b_{y} \  c_{y}]^T$. Similar to the case of in-plane translations, we can collect pixel shifts from multiple locations $(x^{i},y^{i})$ (for $i=1,2,..,A$) in the image to form a overdetermined set of linear equations in terms of the unknowns $\textbf{b}_{x}$ and $\textbf{b}_{y}$ as follows.
\noindent\begin{minipage}{.5\linewidth}
\begin{equation}
\left[ {\begin{array}{c}
   \Delta x^{1} \\
   \Delta x^{2} \\
   .  \\
   \Delta x^{i} \\
\end{array} } \right]
=
\left[ {\begin{array}{ccc}
    x^{1} & y^{1} & 1 \\
    x^{2} & y^{2} & 1 \\
    . & . & . \\
    x^{A} & y^{A} & 1 \\
\end{array} } \right]
\left[ {\begin{array}{c}
    a_{x} \\
    b_{x}  \\
    c_{x}
\end{array} } \right]
\label{abc x_eqn}
\end{equation}
\end{minipage}%
\begin{minipage}{.5\linewidth}
\begin{equation}
\left[ {\begin{array}{c}
   \Delta y^{1} \\
   \Delta y^{2} \\
   .  \\
   \Delta y^{A} \\
\end{array} } \right]
=
\left[ {\begin{array}{ccc}
    x^{1} & y^{1} & 1 \\
    x^{2} & y^{2} & 1 \\
    . & . & . \\
    x^{A} & y^{A} & 1 \\
\end{array} } \right]
\left[ {\begin{array}{c}
    a_y \\
    b_y \\
    c_y
\end{array} } \right]
\label{abc y_eqn}
\end{equation}
\end{minipage}

We use the difference between the extreme points of the locally estimated PSFs to compute the quantities $\Delta x^{i}$ and $\Delta y^{i}$. By making use of multiple PSFs computed from different locations in the blurred image, we can solve for $\textbf{b}_{x}$ and $\textbf{b}_{y}$ using least squares error minimization. From the estimates of $\textbf{b}_{x}$ and $\textbf{b}_{y}$, and with the help of the relations in Eq. \eqref{Delta x eqn} and Eq. \eqref{Delta y_eqn}, the estimated parameters and the normals are related as
\begin{equation}
n_X/(\nu \ n_Z) = a_x/c_x
\end{equation}
\begin{equation}
n_Y/(\nu \ n_Z) = b_y/c_y
\end{equation}
Hence, we can obtain the components of the surface normal (upto a scale factor ambiguity) as follows
\begin{equation}
 n_X :n_Y:n_Z= \nu \  a_x/c_x : \nu \  b_y/c_y : 1
\end{equation}
The common scale factor can be removed by enforcing unit norm constraint to yield the final estimate of normal. Note that the normal estimate obtained in this way not only handles practically occurring camera motion, but also provides a normal estimate with minimal correspondence requirements. A minimum of 3 correspondences is sufficient to obtain the normal estimate by solving Eqs. \eqref{abc x_eqn}-\eqref{abc y_eqn}.

\section{Multi-Planar Motion deblurring}
\label{dblr_multi_pn}
In this section, we introduce our approach for recovery of complete scene geometry and restoration of the latent image assuming the availability of a single motion blurred image of a multi-planar scene. From a single motion blurred image, we can recover the normals corresponding to all the planes. However, to recover the latent image, we need to solve for the remaining unknown variables in the image formation model.

Consider the discrete equivalent model of blurred image formation  as given by
\begin{equation}
\label{gi_defn_disc_eqn}
g(\mathbf{x})=\sum \limits _{i=1}^{N} \alpha_i \odot \bigg( \sum \limits _{p \in P} \omega (p) f(\mathbf{H}_{p,i}^{-1}(\mathbf{x}))\ \bigg)
\end{equation}

From Eq. \eqref{gi_defn_disc_eqn} it can be observed that, for latent image estimation, we need to recover the camera motion ($\omega$), depth values ($d_i$ for i=1,..,N), and an accurate estimate of plane segmentation masks ($\alpha_i$ for i=1,..,N). We first employ the inlier blur kernels and the normal estimate obtained from RANSAC to estimate the TSF ($\omega$) and the depth parameters ($d_i$ for i=1,..,N). This is then followed by an alternating minimization scheme where we solve for both the latent image ($f$) and segmentation masks ($\alpha_i$ for i=1,..,N) to yield the final restored image.
\subsection{Estimation of camera motion and depth values}
\label{secn_cam_depth_estn}
To estimate TSF and depth values, we make use of the inlier PSFs and the normal estimates obtained from RANSAC. Consider a spatial location $\textbf{x}$ lying on the $i^{th}$ plane of the scene. The PSF at $\textbf{x}_j$ can be related to TSF $\omega$ as \cite{paramanand2014shape}
\begin{equation}
\label{psf_tsf_reln}
k(\mathbf{x}_j,\mathbf{u})= \sum \limits _{p \in P} \omega (p) \delta (\mathbf{u} - (\mathbf{H}_{p,i}\mathbf{x}_j-\mathbf{x}_j))
\end{equation}
Eq. \eqref{psf_tsf_reln} relates the inlier PSFs with corresponding depth values and underlying camera motion. Although the camera motion is the same for the entire image, the effective pixel motion experienced by each scene point depends on the normal and the depth value of the corresponding plane. To solve for the TSF, we define the depth of one plane to be reference depth $d_0$ and solve for the scalar factor $ s_i= \frac{d_0}{d_i}$ corresponding to all other planes \cite{paramanand2014shape}.

The relation in Eq. \eqref{psf_tsf_reln} can be expressed in matrix-vector multiplication form as  
\begin{equation}
\label{psf_tsf_reln_mtx_vctr}
\textbf{k}_{\textbf{x}_j}= \textbf{M} _{\textbf{x}_j} \boldsymbol \omega
\end{equation}
where $\textbf{M} _{\textbf{x}_j}$ is a motion matrix which embeds the motion of a point light source at $\textbf{x}_j$ with respect to the camera poses in $P$, and $\textbf{k}_{\textbf{x}_j}$ and $\boldsymbol \omega$ are the column vector forms of $k(\textbf{x}_j)$ and $\omega$. Note that the entries of $\textbf{M}_{\textbf{x}_j}$ depend on the plane normal and unknown scale factor $s_{i}$ too. By aggregating such relations corresponding to all the inlier PSFs we can obtain an equation of the following form.
\begin{equation}
\label{all_psf_tsf_reln_mtx_vctr}
\textbf{k}= \textbf{M} \boldsymbol \omega
\end{equation}
where $\textbf{k} = \left[ {\begin{array}{ccc}
\textbf{k}_{\textbf{x}_1}^T \ .. \ \textbf{k}_{\textbf{x}_c}^T
\end{array} } \right]^T$ and $\textbf{M} = \left[ {\begin{array}{ccc}
\textbf{M}_{\textbf{x}_1}^T \ .. \ \textbf{M}_{\textbf{x}_c}^T
\end{array} } \right]^T$. The total number of inlier PSFs obtained from RANSAC is denoted as $c$. Since the measurement matrix requires knowledge of depth values corresponding to each plane, we cannot use Eq. \eqref{all_psf_tsf_reln_mtx_vctr} alone to solve for the camera motion $\omega$. Hence we choose to alternatively update the camera motion $\omega$ and depth values until convergence.

\noindent \textbf{TSF refinement: }
Once the scale factors are known, we can build the matrix $\textbf{M}$ in Eq. \eqref{all_psf_tsf_reln_mtx_vctr} and then estimate $w$ by solving the following optimization problem
\begin{equation}
\hat{\boldsymbol \omega}^m = \min_{\boldsymbol \omega}  \parallel \textbf{k}-\textbf{M} \boldsymbol \omega \parallel _2^2 + \lambda_{\boldsymbol \omega} \parallel \boldsymbol \omega \parallel_1,
\label{eqn_opt_w}
\end{equation}
where $m$ denotes the iteration number. We apply an $L_1$ norm based sparsity prior on $\omega$ to enforce the fact that camera motion will occupy only few poses in the entire search space. The weight of the prior is controlled through the scale factor $\lambda_{\boldsymbol \omega}$. We solve Eq. \eqref{eqn_opt_w} using alternating  direction  method  of  multipliers (ADMM) \cite{boyd2011distributed} to obtain the TSF estimate $\hat{\boldsymbol \omega}^m$ for $m^{th}$ iteration.
 
\noindent \textbf{Scale factor refinement:}
To refine the scale factors, we form a set of scale factors $S$ around 1, and search for the ones which satisfy the current estimate of TSFs and the inlier PSFs corresponding to each plane. We first use the camera motion $\hat{\boldsymbol \omega}^m$ obtained from previous iteration to generate the PSFs at all the locations and all the scale factors in $S$. For $i^{th}$ plane, the kernels generated from $\hat{\boldsymbol \omega}^m$ at locations corresponding to all the inlier PSFs of that plane are compared with respective inlier PSFs to update the scale factor $s_{i}$. To update $s_{i}$ we solve the following optimization problem
 \begin{equation}
s_{i}^{m} = \min_{s \in S} \sum \limits_{\textbf{x}_j \in X_i} \parallel \textbf{k}_{\textbf{x}_j} - \textbf{M} _{(\textbf{x}_j,s)} \hat{\boldsymbol \omega}^m \parallel_2^2 \quad \text{for } i=1,2,..., N
\label{eqn_opt_s}
\end{equation}
where $X_i$ refers to the set of spatial locations corresponding to all the inlier PSFs of $i^{th}$ plane.

In the first iteration, we estimate the TSF by setting all the scale factors to unity (i.e; $s_{i}=1$ $\forall i$). The TSF estimate thus obtained is then used for updating the scale factors. Using the updated scale factors, we re-estimate $w$ using \eqref{eqn_opt_w}. This refinement process of $w$ and $s_i$ is repeated until the convergence of all the scale factors.
\subsection{Image restoration and recovery of segmentation masks}
In this section, we will discuss our approach to recover the latent image by making use of the estimates obtained from previous sections. Since latent image estimation requires knowledge of segmentation masks, the problem is still ill-posed. Hence we employ an alternating minimization (AM) scheme, where we iteratively repeat both latent image estimation and segmentation mask recovery to arrive at the desired solution. Details on the two sub-problems in our AM scheme is discussed next.

\noindent \textbf{Latent image estimation:}
The relation in Eq. \eqref{gi_defn_disc_eqn} can be expressed in a matrix-vector multiplication form as follows
\begin{equation}
\label{gi_defn_disc_eqn2}
\textbf{g}=\sum \limits _{i=1}^{N} \boldsymbol \Gamma _{\alpha_{i}} \textbf{W}_i \textbf{f} = \textbf{W} \textbf{f}
\end{equation}
where $\textbf{g}$ and $\textbf{f}$ are the lexicographically ordered form of $g$ and $f$ ,respectively. The matrix $\textbf{W}_i$ which embeds the pixel motion corresponding to $i^{th}$ plane is built according to the camera motion $\omega$ and the parameters of $i^{th}$ plane. $\boldsymbol \Gamma _{\alpha_{i}}$ is a diagonal matrix built based on the segmentation mask $\alpha _i$. The matrix $\textbf{W} = \sum \limits _{i=1}^{N} \boldsymbol \Gamma _{\alpha_{i}} \textbf{W}_i$ subsumes the pixel motions corresponding to all the points in the scene. From known estimates of the scene plane parameters ($n_i$, $d_i$), camera motion ($\omega$), and plane segmentation masks ($\alpha_i$), we estimate the latent image $f$ by solving the following form of optimization.
\begin{equation}
\label{lat_img_estn}
\widehat{f}=\min \limits_{f} \parallel \textbf{W} \textbf{f} - \textbf{g} \parallel_2^2 + \lambda _f \parallel \triangledown \textbf{f} \parallel _1
\end{equation}
Here, to obtain $\widehat{f}$, we apply $L_1$ norm based prior (weighted by the scale factor $\lambda _f$) to enforce natural sparsity of latent image gradients \cite{wang2008new} and then solve the resulting optimization using ADMM \cite{boyd2011distributed}.

\noindent \textbf{Estimation of segmentation masks:} We estimate segmentation masks by posing it as a multi-label MRF optimization problem where the labels indicating the pixel assignments corresponding to each plane. This optimization is then solved using graphcut \cite{boykov2001fast}. For a pixel at $\textbf{p}$, we define the cost corresponding to assigning the label $l_\textbf{p}$ as
\begin{equation}
C(l_\textbf{p})= DC(l_\textbf{p}) + \lambda _l \sum _{\textbf{q} \in \mathcal{N} _\textbf{p}} SC _{\textbf{p}, \textbf{q}} (l _\textbf{p}, l _\textbf{q})
\label{eq12}
\end{equation}
where $DC(l_\textbf{p})$ is the data cost to assign the label $l_\textbf{p}$ to pixel $\textbf{p}$, $\mathcal{N} _\textbf{p}$ is a neighborhood of pixels around $\textbf{p}$, $SC _{\textbf{p}, \textbf{q}}(l _\textbf{p}, l _\textbf{q})$ is the smoothness cost to assign the labels $(l _\textbf{p}, l _\textbf{q})$ to the adjacent pixels $\textbf{p}$, $\textbf{q}$ and $\lambda _l$ is the scalar weight on the smoothness term. We use the following form of cost function to compute the data cost corresponding to $l_\textbf{p} = i$.
\begin{equation}
\label{eqn_plane_datacost}
DC ( l_\textbf{p} = i ) = \parallel \textbf{g} - \textbf{W}_i \textbf{f} \parallel _2^2
\end{equation}
It is straightforward to see that the above data cost enforces the label assignment to respect the image formation model in Eq. \eqref{gi_defn_disc_eqn2}. The smoothness cost $SC _{\textbf{p}, \textbf{q}} (l _\textbf{p}, l _\textbf{q})$ has the following form.
\begin{equation}
\label{eqn_smothnesscost}
SC _{\textbf{p}, \textbf{q}} (l _\textbf{p}, l _\textbf{q}) = 1-r^{|l _\textbf{p} - l _\textbf{q}|}
\end{equation}
where $r$ is a scalar value. This is used to enforce the fact that adjacent pixels in the image are more likely to have identical labels, i.e; the pixels corresponding to a single plane will form a contiguous region.

We start our AM by solving for the latent image by initializing $\alpha_i$ corresponding to the background layer as all $1$s and other layers as all $0$s. This is then followed by alternative refinement of both mask and the latent image to yield the final restored image as well as an accurate layer segmentation map.

\section{Experiments}
In this section, we validate the proposed method on both synthetic and real examples. We also show quantitative and qualitative comparisons with state-of-the-art blind deblurring approaches. For normal estimation of all the scene planes, we have estimated the PSFs from overlapping patches of size $120 \times 120$ with an overlap factor of $50$. To estimate the blur kernel for a selected patch we used an off-the-shelf blind motion deblurring technique in \cite{xu2010two}. To find the extremities of blur kernels, we use the PSF end point localization approach from \cite{zheng2011structure}. These PSF estimates are then used in our RANSAC \cite{fischler1981random} based approach to identify the number of planes and associated normals. In the RANSAC algorithm, the PSF estimate which induces a deviation of more than $11$ degrees in the normal estimate is treated as an outlier.

In all our experiments, the number of iterations for alternating refinement of TSF and depth values in Section \ref{secn_cam_depth_estn} as well as the AM between the latent image estimation and segmentation mask recovery was set to $5$. To solve various optimization problems discussed in previous sections, the value of $\lambda _{\boldsymbol \omega}$, $\lambda _{f}$, and $r$ were set to $0.1$, $0.002$, and $0.8$, respectively. All these parameters were found empirically through experimentation. For the  segmentation mask recovery using Eq. \eqref{eq12}, we used the image obtained by applying $L_0$ smoothing filter \cite{xu2011image} on the estimated latent image from Eq. \eqref{lat_img_estn}. As observed in \cite{hu2014joint}, the $L_0$ smoothing filter not only helps in countering the adverse effects of the small edges during depth estimation, it also helps in recovering strong gradients in the latent image which, in turn, ensure better convergence of the subsequent AM approach.

\begin{figure*}[t]
\centering
\begin{tabular}{cccccc}
\includegraphics[width=.22\linewidth]{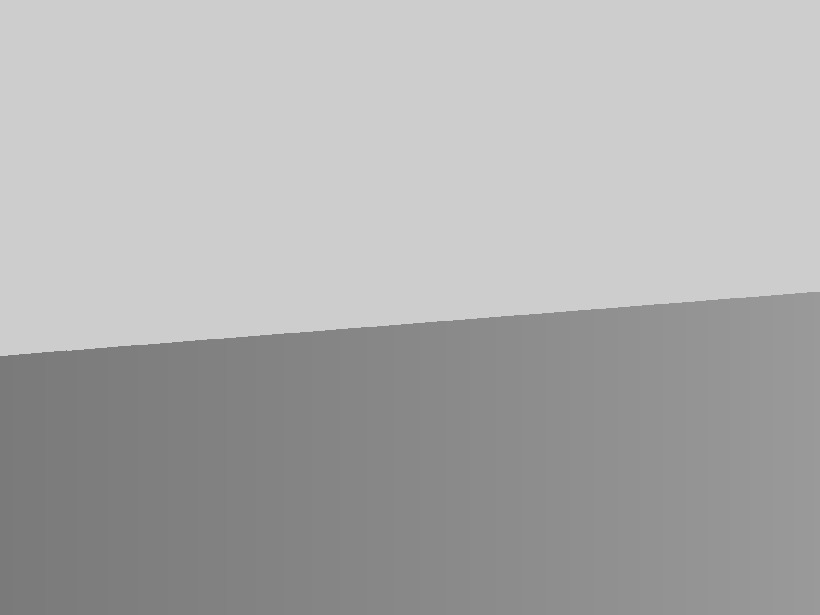}&
\includegraphics[width=.22\linewidth]{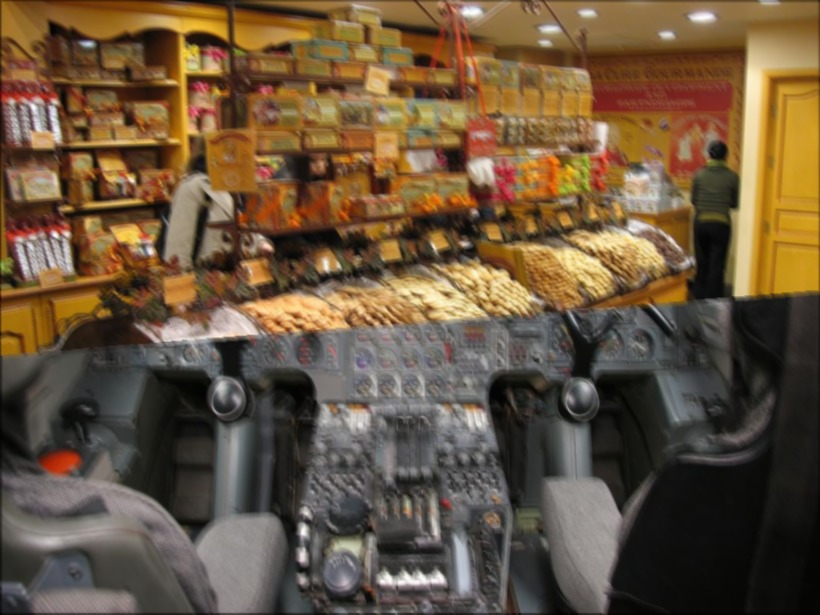}&
\includegraphics[width=.22\linewidth]{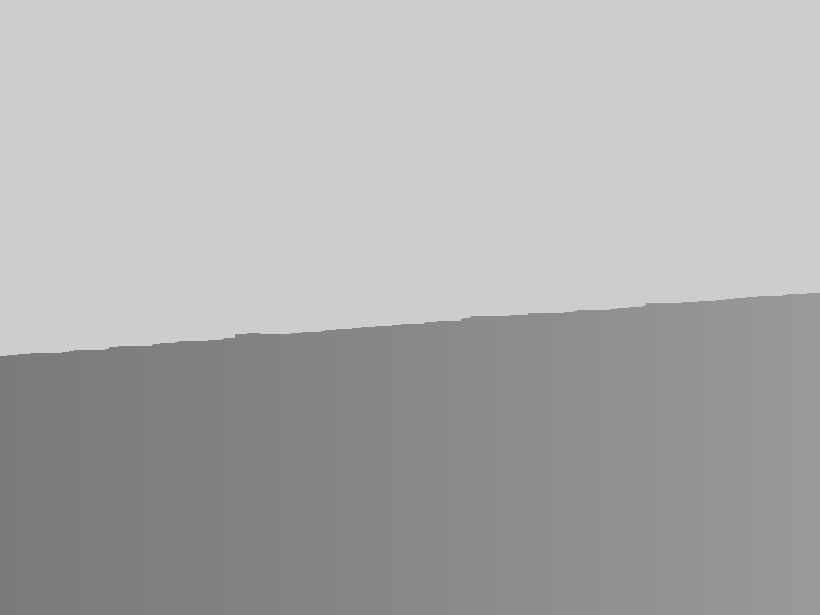}&
\includegraphics[width=.22\linewidth]{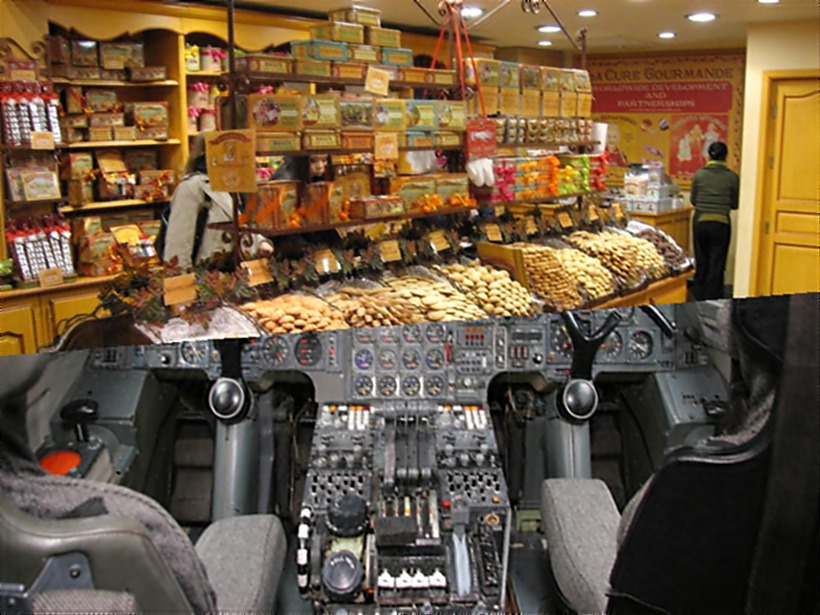}\\
(a)&(b)&(c)&(d)\\
\includegraphics[width=.22\linewidth]{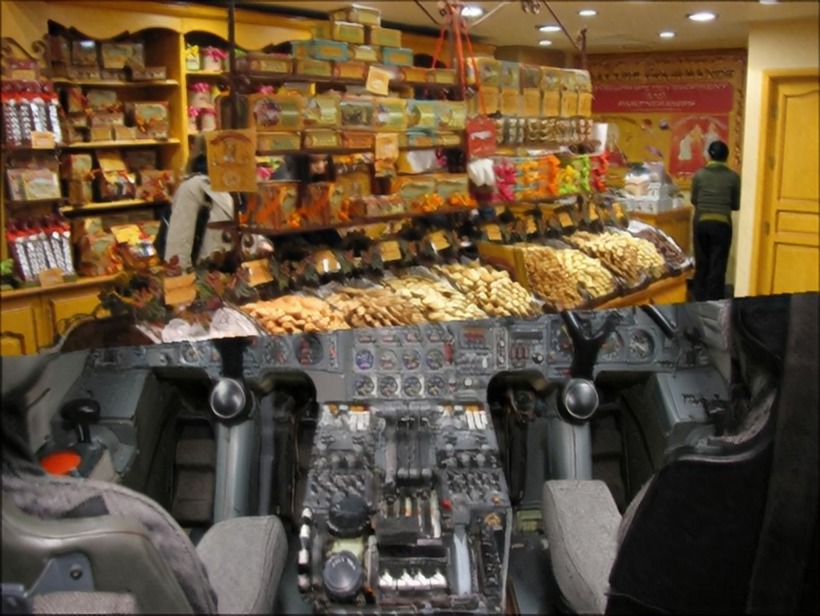}&
\includegraphics[width=.22\linewidth]{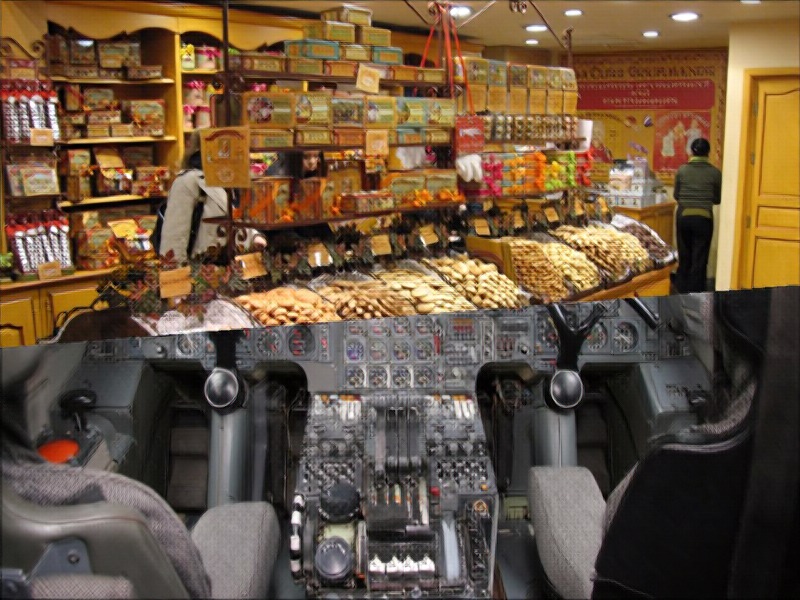}&
\includegraphics[width=.22\linewidth]{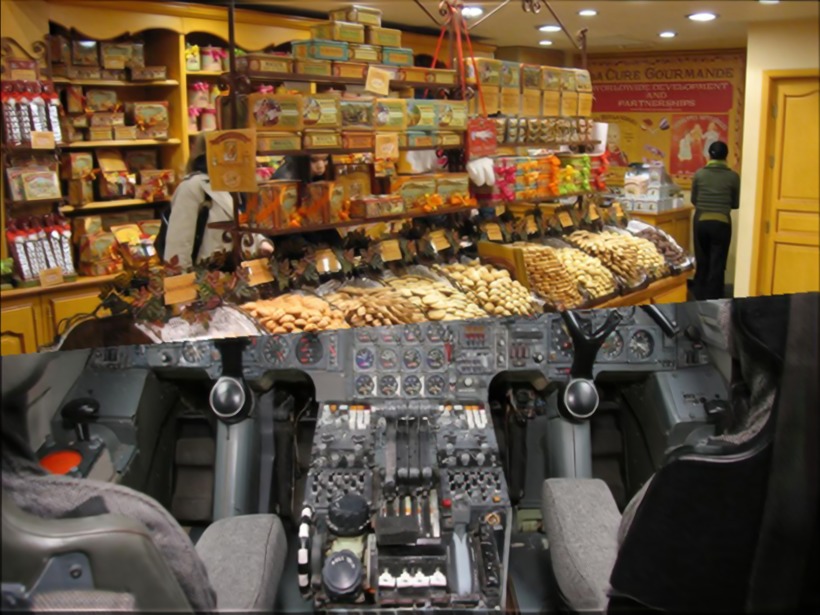}&
\includegraphics[width=.22\linewidth]{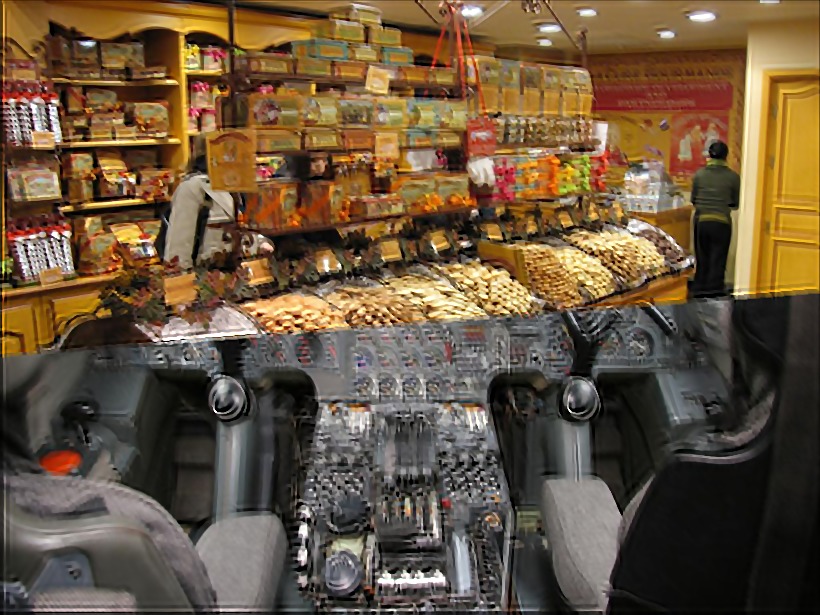}
\\
(e)&(f)&(g)&(h)\\
\end{tabular}
   \caption{Results for a synthetically blurred two-layer scene, with background layer as a fronto-parallel plane and foreground layer having $\textbf{n} = [0 \ -0.3162 \   0.9701]^T$. (a) Ground truth depth map (generated using the plane parameters and segmentation masks). (b) Input blurred image generated using the depth map and camera trajectory from \cite{kohler2012recording}. (c) Recovered depth-map obtained using the estimated plane parameters and segmentation masks. Restored image using (d) the proposed approach, (e) \cite{nah2017deep} , (f) \cite{kupyn2018deblurgan}, (g) \cite{tao2018scale} and (h) \cite{hu2014joint}.}
\label{synth2}
\end{figure*}

\begin{figure*}[htbp]
\centering
\begin{tabular}{cccccc}
\includegraphics[width=.22\linewidth]{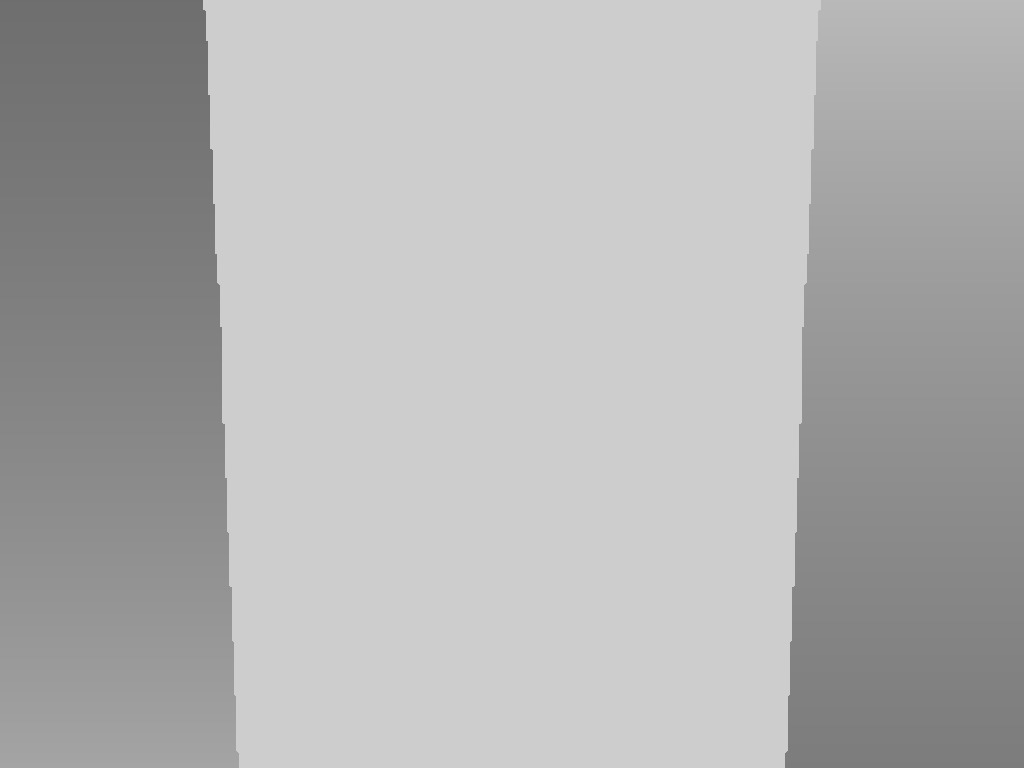}&
\includegraphics[width=.22\linewidth]{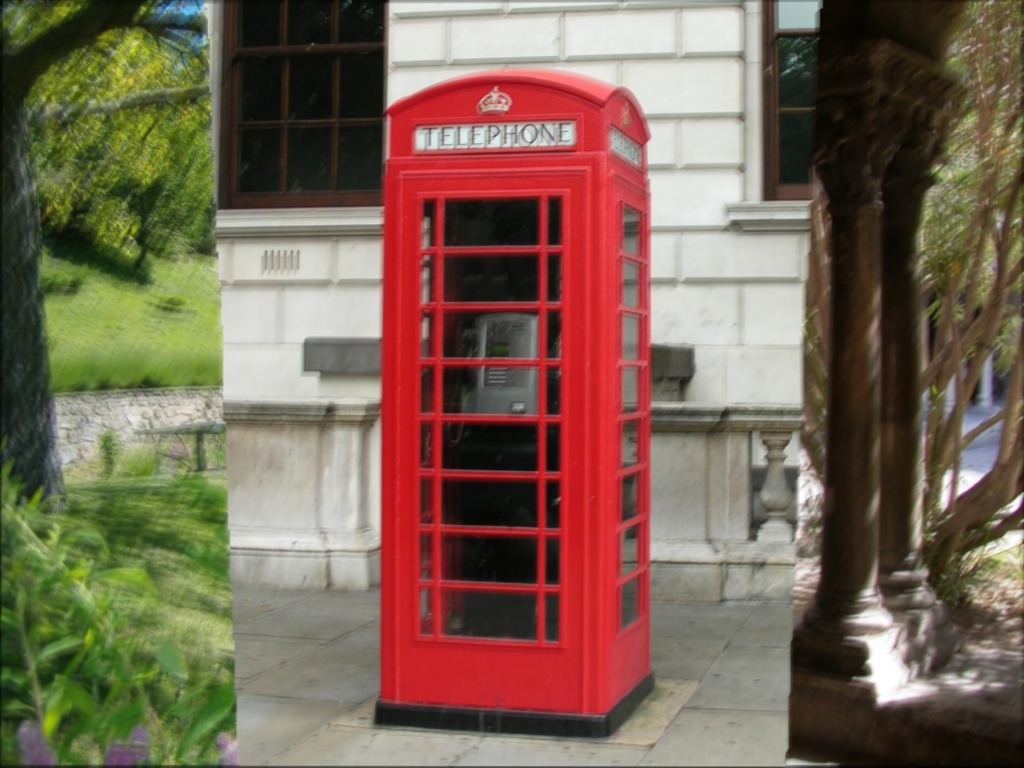}&
\includegraphics[width=.22\linewidth]{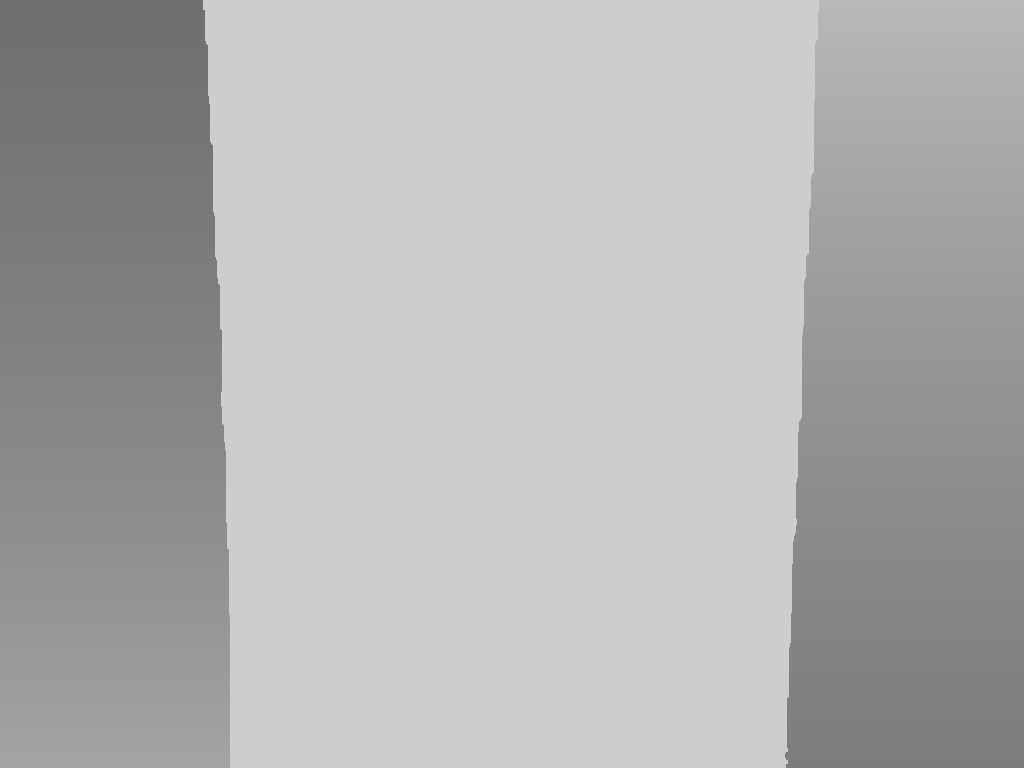}&
\includegraphics[width=.22\linewidth]{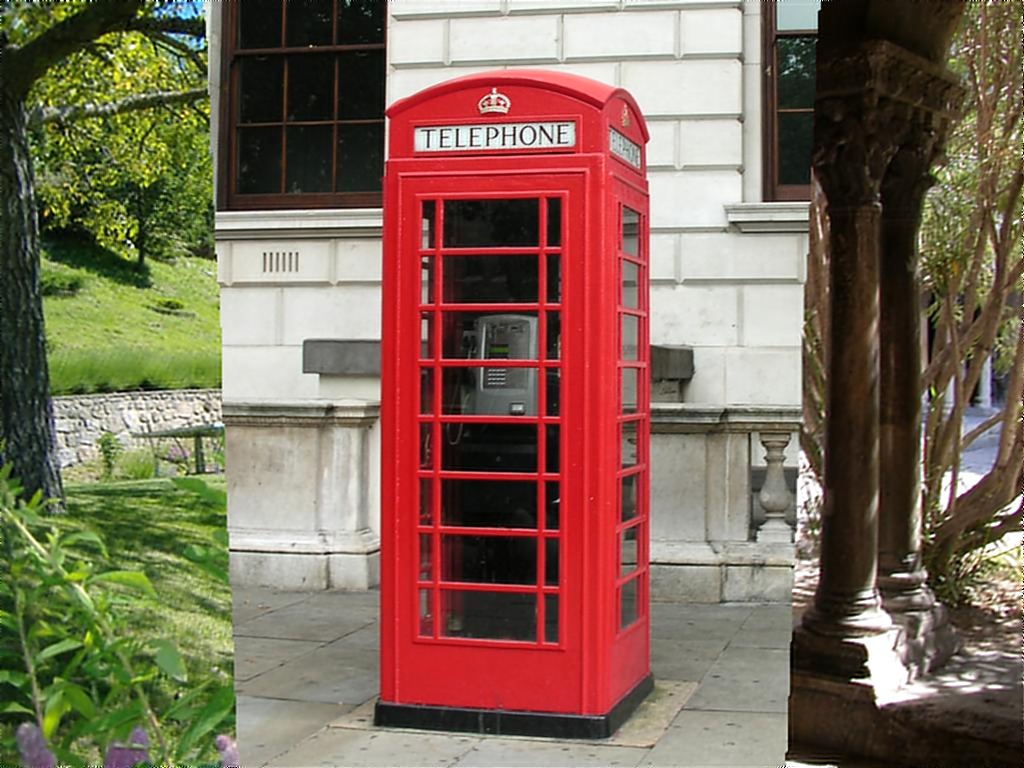}\\
(a)&(b)&(c)&(d)\\
\includegraphics[width=.22\linewidth]{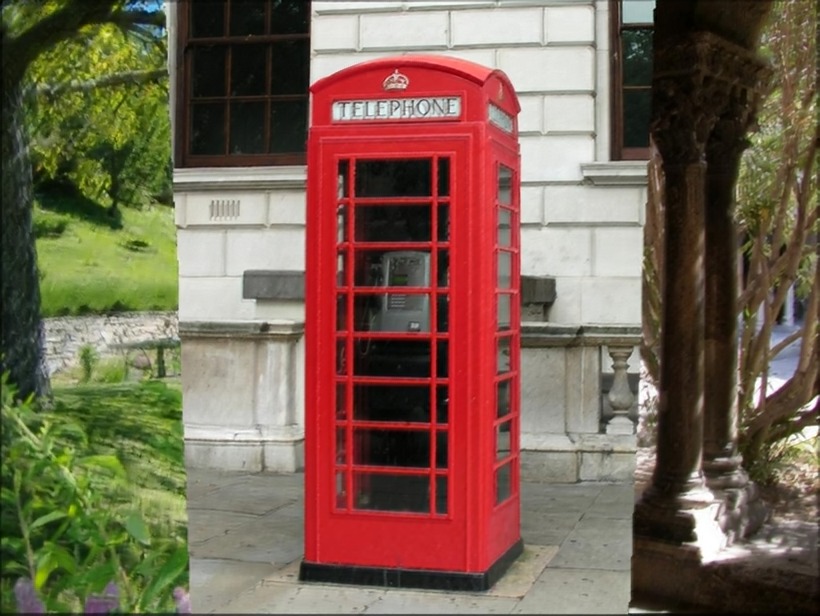}&
\includegraphics[width=.22\linewidth]{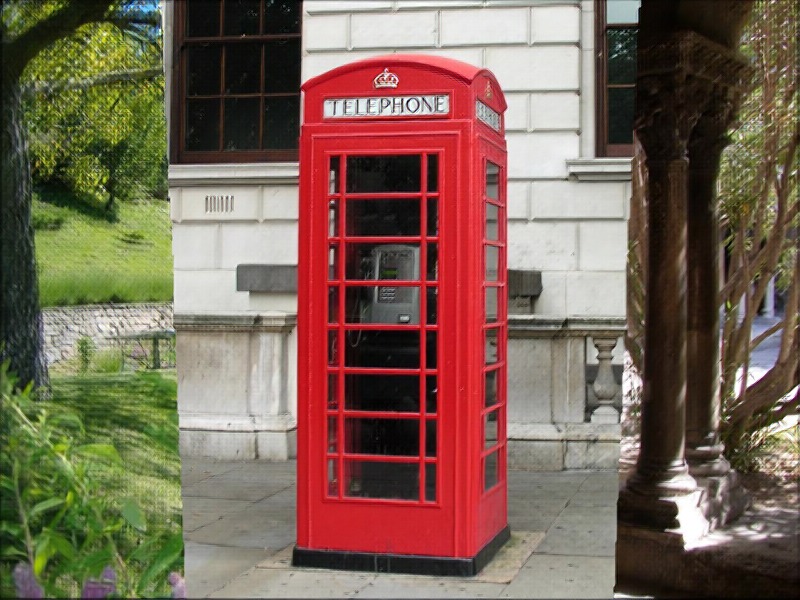}&
\includegraphics[width=.22\linewidth]{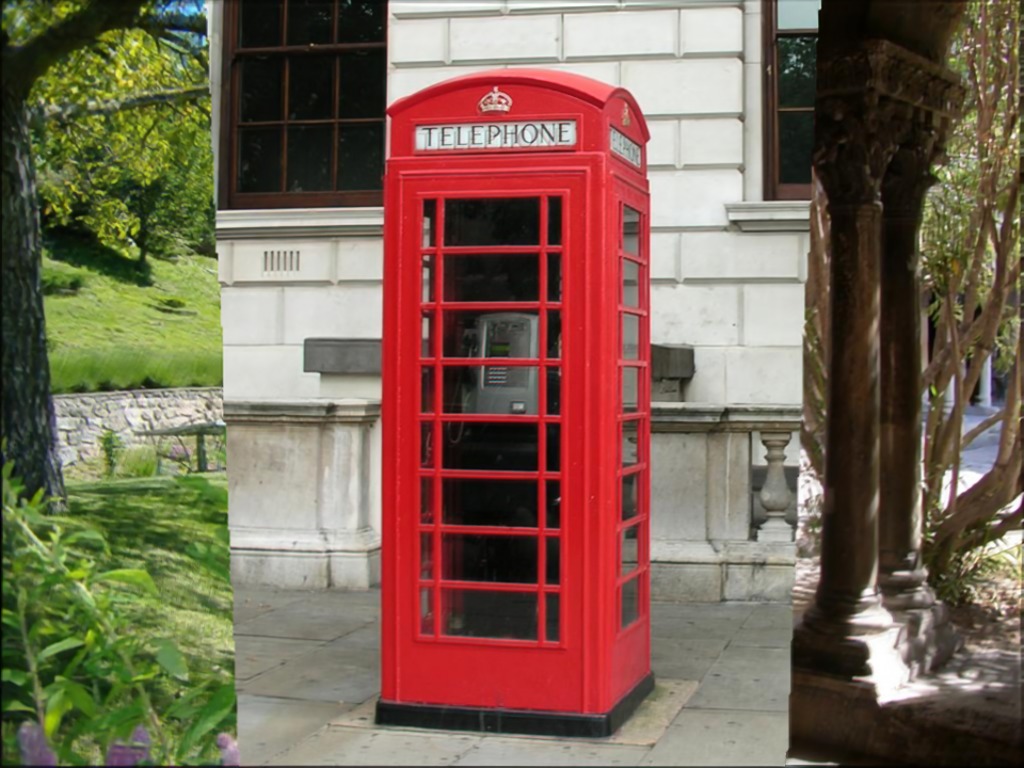}&
\includegraphics[width=.22\linewidth]{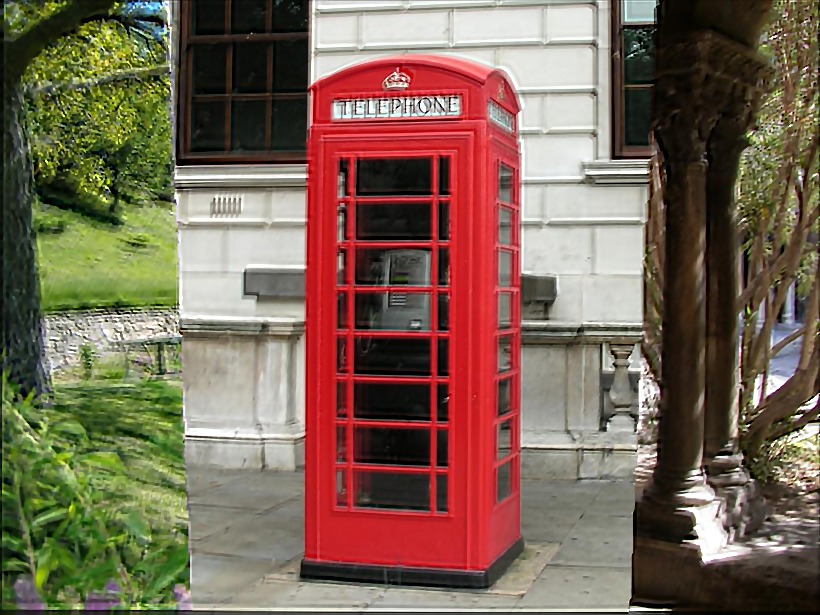}
\\
(e)&(f)&(g)&(h)\\
\end{tabular}
   \caption{Results for a synthetically blurred three-layer scene. (a) Ground truth depth map (generated using the plane parameters and segmentation masks). (b) Input blurred image generated using the depth map and camera trajectory from \cite{kohler2012recording}. (c) Recovered depth-map obtained using the estimated plane parameters and segmentation masks. Restored image using (d) the proposed approach, (e) \cite{nah2017deep} , (f) \cite{kupyn2018deblurgan}, (g) \cite{tao2018scale} and (h) \cite{hu2014joint}.}
\label{synth3}
\end{figure*}

\subsection{Synthetic Experiments}
To generate synthetic test examples, we used the trajectories from the dataset of \cite{kohler2012recording} to simulate the camera motion. Images from the data-set of \cite{Sun:2012:SR_sun_hays} were used as ground truth images corresponding to different layers. Layer masks were formed by manually creating binary masks of arbitrary shapes. For all the synthetic experiments we set the focal length to be 1000 pixels.

To perform quantitative evaluation of our proposed scheme for normal and latent image estimation, we created a dataset of synthetic examples comprising of 10 blurred images corresponding to 3D scenes with single and multiple planes. We verify the performance of our normal estimation scheme by finding the angular error between the ground truth normal and estimated normal. For the performance comparison of our deblurring scheme, we have used PSNR (Peak Signal to Noise Ratio) and SSIM (Structural Similarity Measure) values of the restored images calculated with respect to the corresponding ground truth images. These values are compared with state-of-the-art deblurring approaches, by generating their results using the implementations provided by respective authors.

 Fig.~\ref{synth2} and Fig.~\ref{synth3} show synthetic examples corresponding to a $2$ and a $3$ layer scenes, respectively. In both cases, images were blurred using camera motion involving translations and rotations and the background was set to be fronto-parallel. While the foreground layer of example in Fig.~\ref{synth2} was blurred using ${n} = [0 \ -0.3162 \   0.9701]^T$, we used the normals $[0.3162 \ 0 \   0.9487]^T$ and $[-0.3162 \ 0 \   0.9487]^T$ for the two foreground layers in Fig.~\ref{synth3}.  For scene in Fig.~\ref{synth2}, the estimated normals using the proposed method was found to be $[-0.07 \ -0.3533  \  0.8956]^T$ and $[-0.1 \ -0.1  \  0.92]^T$ which amounts to an average error of $7.8$ degrees. Proceeding similarly for Fig.~\ref{synth3}, the average angular error for the three normals was found to be $6.6$ degrees. The average angular error for our synthetic dataset is $8.15$ degrees. 

Qualitative comparisons for deblurring are shown in Figs.\ref{synth2} and \ref{synth3}. It can be seen that our approach recovers scene texture faithfully, while the results of existing methods contain visible artifacts. The learning based approaches \cite{nah2017deep}, \cite{kupyn2018deblurgan} and \cite{tao2018scale} contain artifacts at the planar boundaries and in dense textured regions. Although undesirable, such local deviations from ground-truth are often found in results of generative models, since the outputs of these networks are not constrained to follow the image formation model. The approach of \cite{Yan_2017_CVPR} leads to deblurring of only few regions since it does not model depth variations. Similar issues are found in the results of multi-planar deblurring algorithm of \cite{hu2014joint} in Fig.~\ref{synth2} and \ref{synth3}, as it does not handle inclined planes. Note that manually marked regions (belonging to each plane) were provided as input to \cite{hu2014joint}). In contrast, our method is able to automatically segment the scenes and deblur them faithfully. The superiority of our results is also reflected in the quantitative comparisons provided in Table \ref{Table1}.

\begin{table}[htbp]
\centering
\caption{\bf Quantitative Comparison of deblurring using our method with other state-of-the-art blind deblurring algorithms on synthetically blurred dataset.}
\begin{tabular}{ccccccc}
\hline
Method & \cite{nah2017deep} & \cite{kupyn2018deblurgan} &  \cite{tao2018scale}   & \cite{hu2014joint} & Ours \\
\hline
PSNR(dB) & 25.49 & 25.23 & 26.02 & 27.25 & 29.12  \\
SSIM  & 0.7200 & 0.7573 & 0.7783 & 0.8346 & 0.9068 \\
\hline
\end{tabular}
\label{Table1}
\end{table}

\subsection{Real Experiments}

\begin{figure*}[htbp]
\centering
\begin{tabular}{ccccccc}
\includegraphics[width=.18\linewidth]{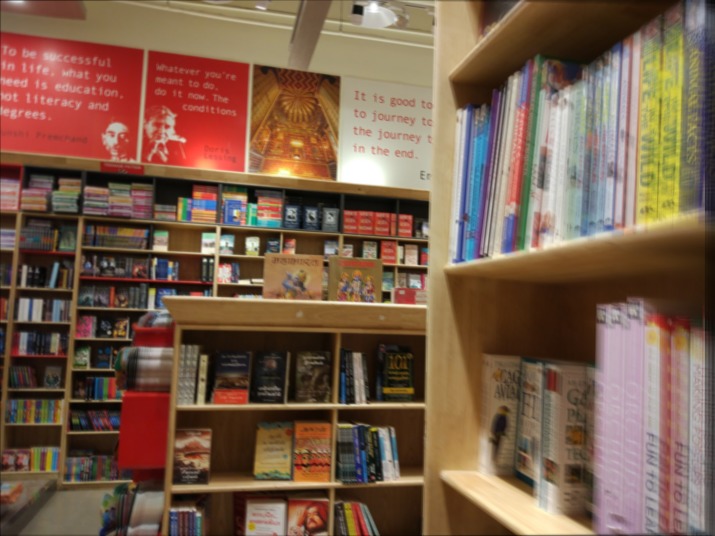}&
\includegraphics[width=.18\linewidth]{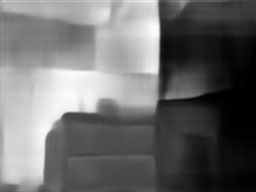}&
\includegraphics[width=.18\linewidth]{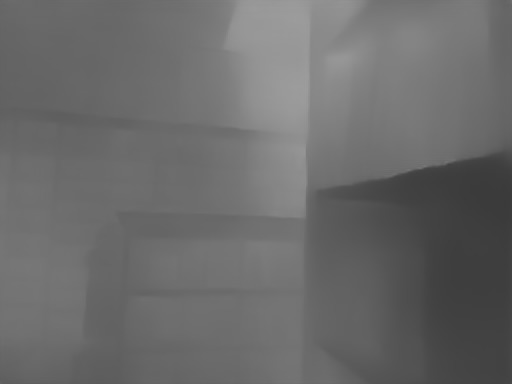}&
\includegraphics[width=.18\linewidth]{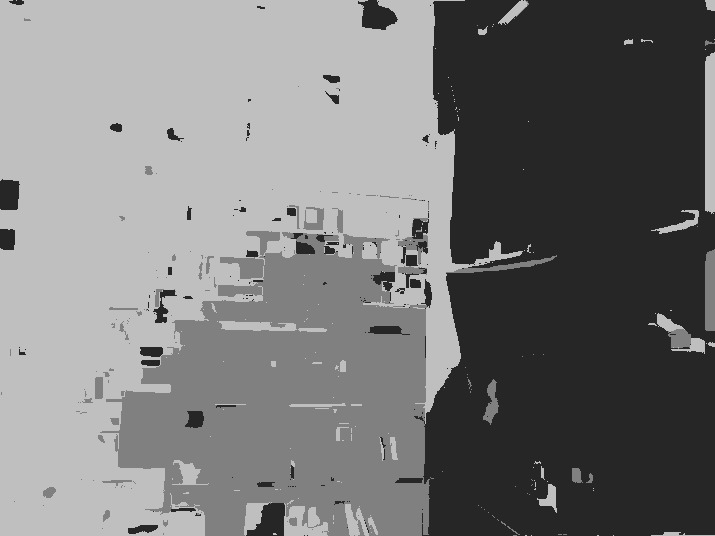}&
\includegraphics[width=.18\linewidth]{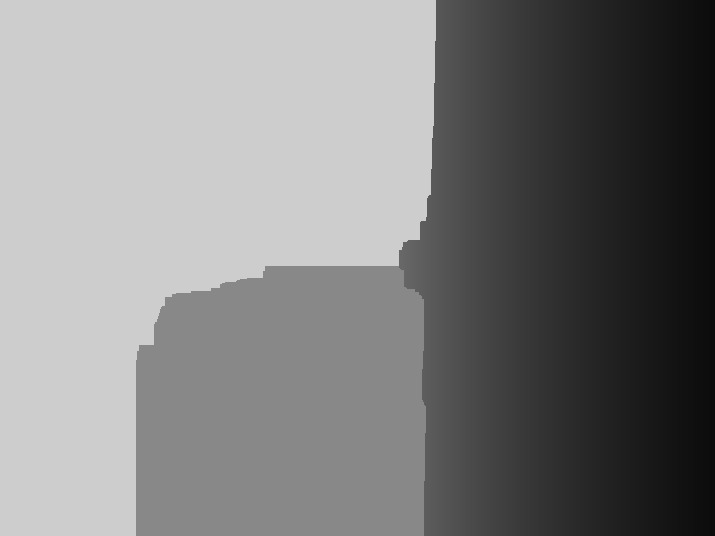}\\
(a)&(b)&(c)&(d)&(e)\\
\includegraphics[width=.18\linewidth]{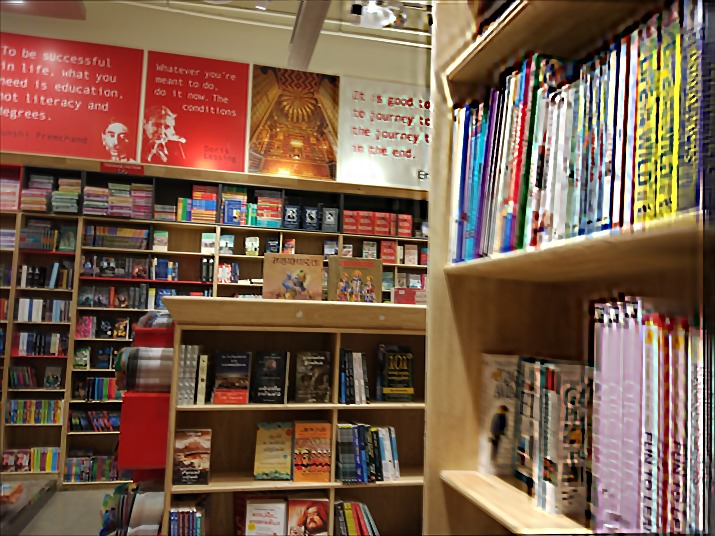}&
\includegraphics[width=.18\linewidth]{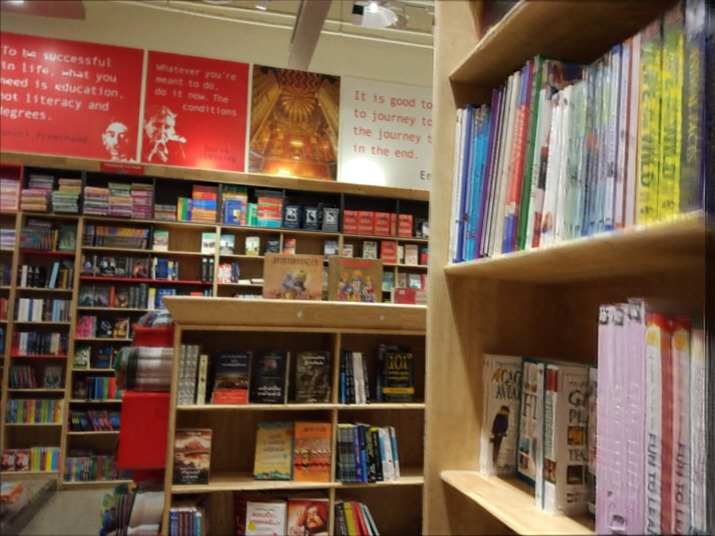}&
\includegraphics[width=.18\linewidth]{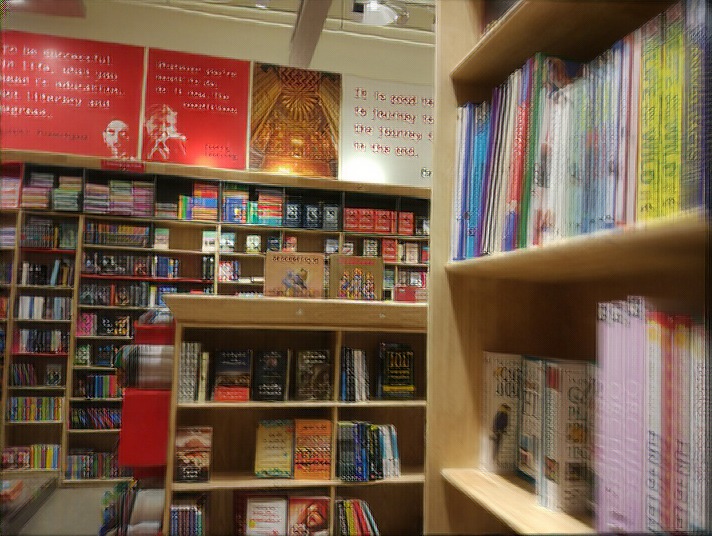}&
\includegraphics[width=.18\linewidth]{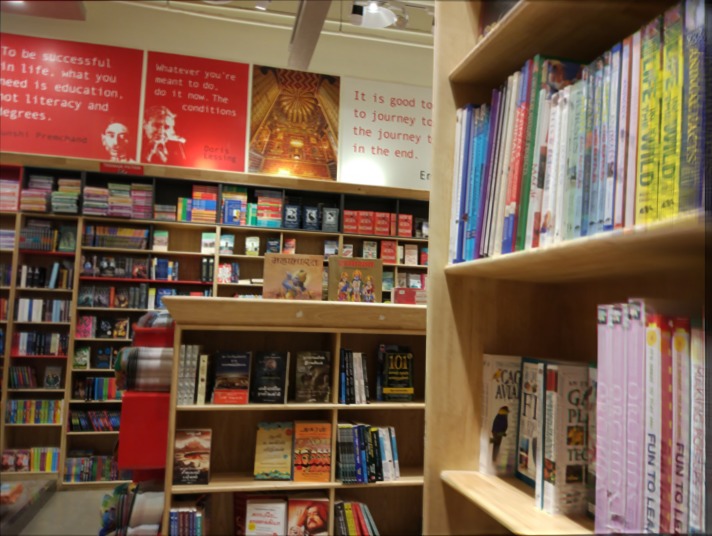}&
\includegraphics[width=.18\linewidth]{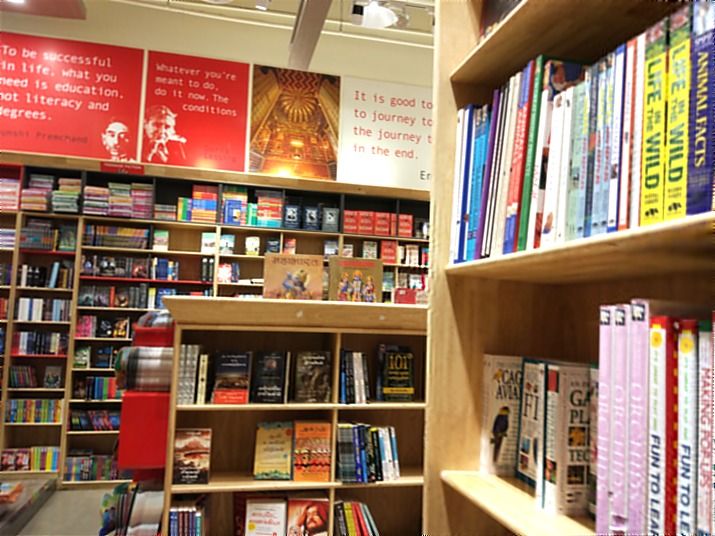}
\\
(f)&(g)&(h)&(i)&(j)\\
\includegraphics[width=.18\linewidth]{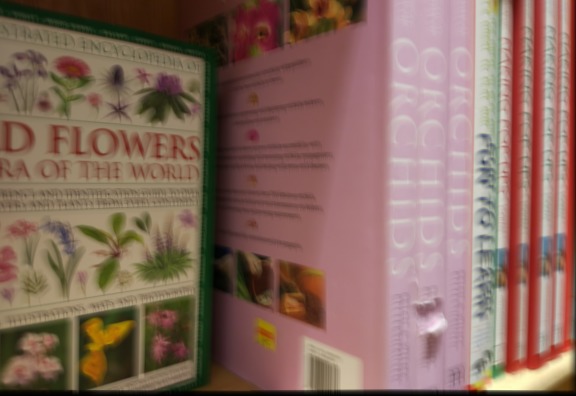}&
\includegraphics[width=.18\linewidth]{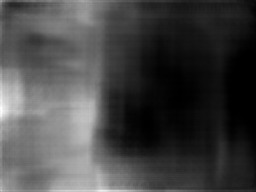}&
\includegraphics[width=.18\linewidth]{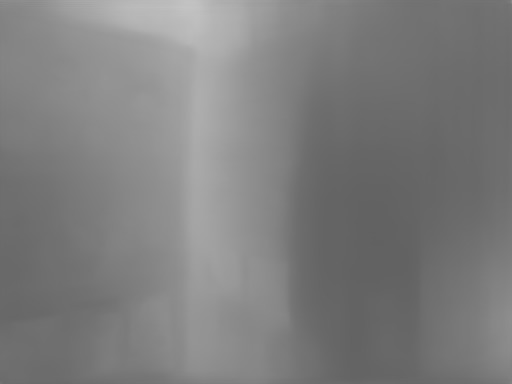}&
\includegraphics[width=.18\linewidth]{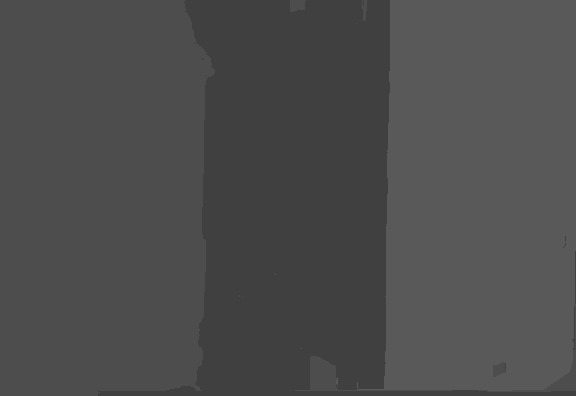}&
\includegraphics[width=.18\linewidth]{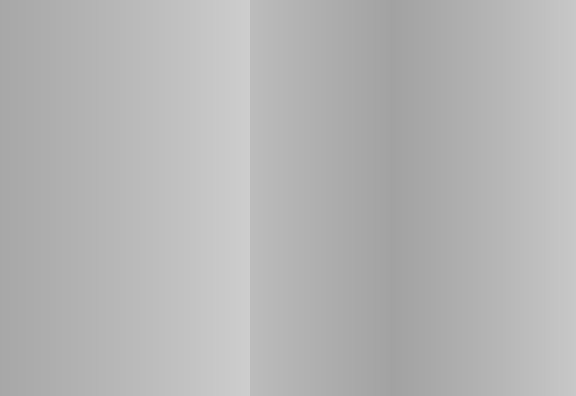}\\
(k)&(l)&(m)&(n)&(o)\\
\includegraphics[width=.18\linewidth]{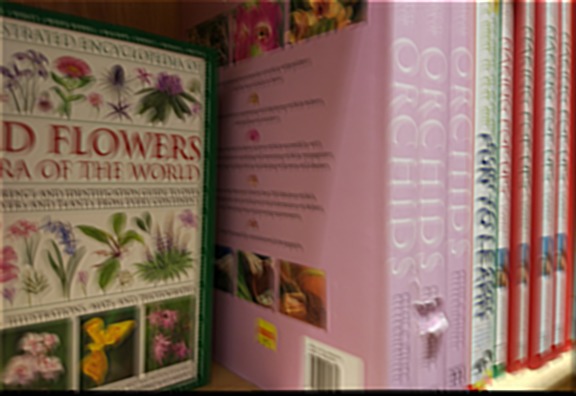}&
\includegraphics[width=.18\linewidth]{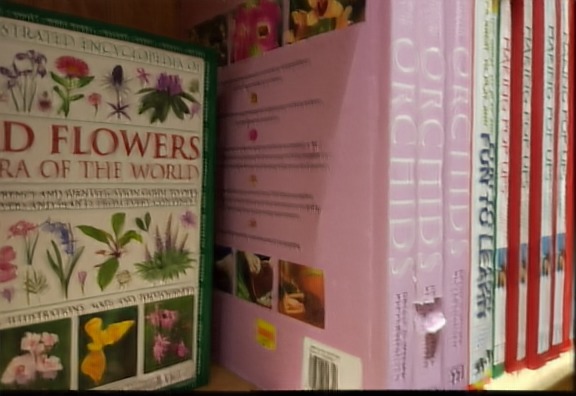}&
\includegraphics[width=.18\linewidth]{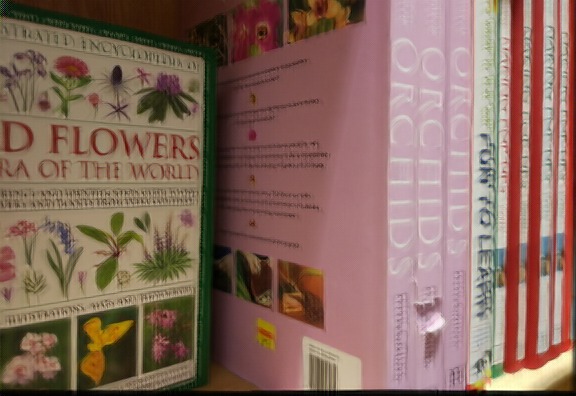}&
\includegraphics[width=.18\linewidth]{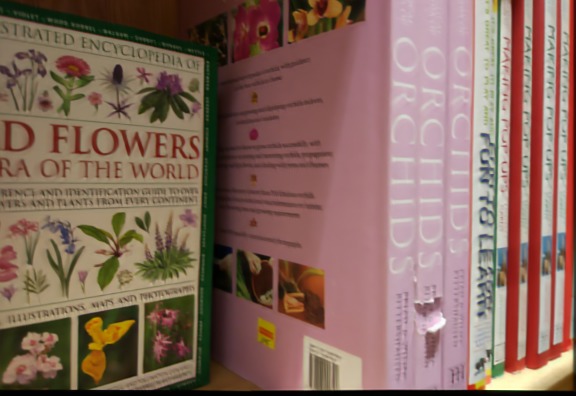}&
\includegraphics[width=.18\linewidth]{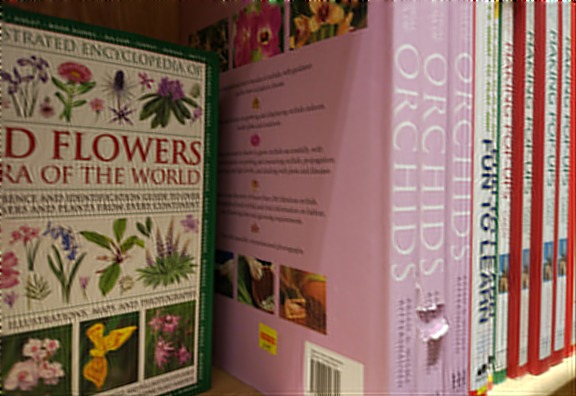}
\\
(p)&(q)&(r)&(s)&(t)\\
\end{tabular}
   \caption{Results of depth estiamtion and deblurring for scenes containing three planes.  Subfigures (a,k) show the input blurred images, (b,l) the depth maps generated using \cite{laina2016deeper}, (c,m) the depth maps generated using \cite{li2018megadepth}, (d,n) The depth-maps used by \cite{hu2014joint}, (e,o) estimated depth map using our method. The second and fourth rows show the deblurring results of \cite{hu2014joint} (f,p), \cite{nah2017deep} (g,q), \cite{kupyn2018deblurgan} (h,r), \cite{tao2018scale} (i,s), and our method (j,t).}
\label{real3}
\end{figure*}

The real experiments are carried out using images captured with Xiomi Mi5 camera in the presence of general camera shake. For the purpose of comparison of deblurring, we applied the conventional non-uniform motion deblurring method of \cite{hu2014joint} and the learning based models of \cite{nah2017deep}, \cite{kupyn2018deblurgan}, and \cite{tao2018scale} to individual images. For depth estimation, we compare with the recent learning based single image depth estimation methods of \cite{laina2016deeper} and \cite{li2018megadepth}.

In the first example, we consider a scenario where a large billboard is present at an inclination to the camera. The scene can be modeled as a single inclined plane. By following the same procedure as outlined in the synthetic case, outlier PSFs were removed and only the authentic blur kernels were used to estimate the TSF. Note that for real examples, we do not have knowledge of the true normal. Using our algorithm, the estimated value of normal for this image is $[0.2379   -0.1738    0.9040]^T$ which is visually consistent with the scene inclination. Note that our estimated depth-map appears more consistent with the scene than the results of \cite{laina2016deeper,li2018megadepth}, since we utilize the information present in the blur-kernels and enforce a planar constraint. 

 Our depth-estimates concur with the scene geometry. The results of \cite{laina2016deeper,li2018megadepth} do describe the scene depth-variation at a very coarse-level but contain various depth-discontinuities at fine-level. The superiority of our depth segmentation can be attributed to the constraints present in our deblurring algorithm.

 In terms of deblurring performance, The method of \cite{hu2014joint} is able to partially deblur some regions in the scene (due to the manually supplied depth-segmentation as input), but suffers from incomplete deblurring and ringing artifacts in inclined regions. The results of \cite{nah2017deep}, \cite{kupyn2018deblurgan}, and \cite{tao2018scale} suffer from incomplete deblurring while introducing artifacts in textured regions. Our method leads to better deblurring results.

\begin{figure*}[htbp]
\centering
\begin{tabular}{cccccc}

\includegraphics[width=.15\linewidth]{real_exp/input/3p1.jpg}&
\includegraphics[width=.15\linewidth]{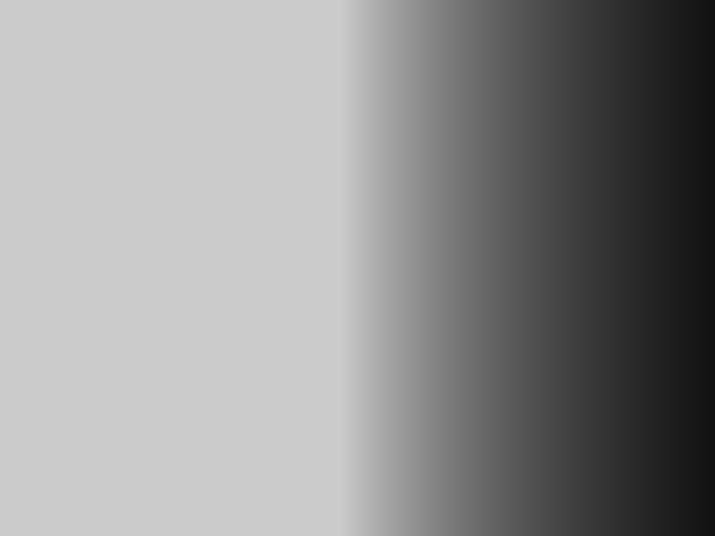}&
\includegraphics[width=.15\linewidth]{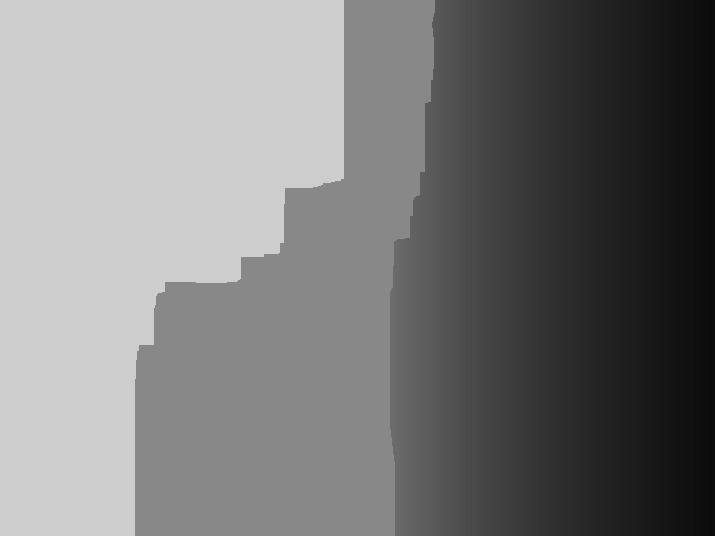}&
\includegraphics[width=.15\linewidth]{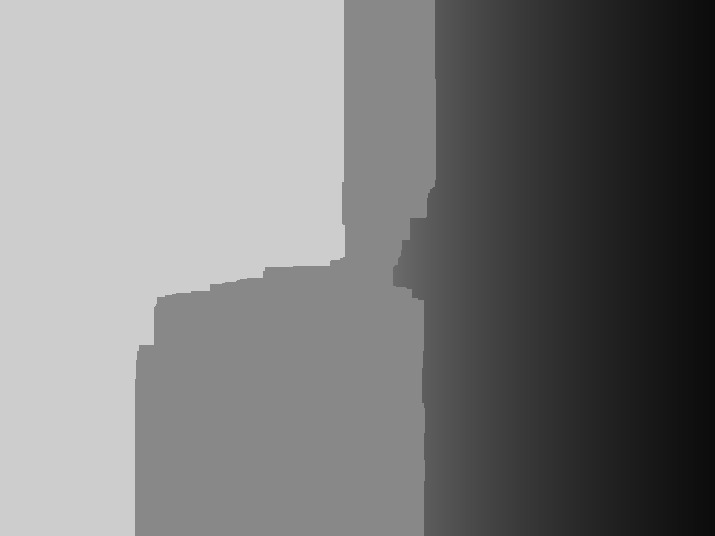}&
\includegraphics[width=.15\linewidth]{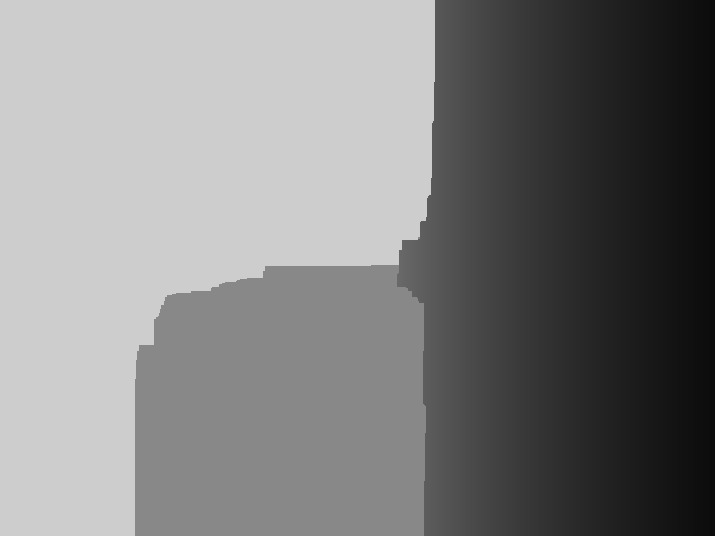}&
\includegraphics[width=.15\linewidth]{real_exp/ours/depth_3p1/depth_GC_5.jpg}\\
(a)&(b)&(c)&(d)&(e)&(f)\\
\end{tabular}
   \caption{Estimated depth-maps for the real blurred image from Fig. \ref{real3}(a) from our AM scheme, recorded after each iteration. Note that the three different planes are clearly distinguishable in the final iteration.}
\label{real4}
\end{figure*}

The next set of examples containing $3$ layered scenes are shown in Fig.~\ref{real3}. The intermediate results for iterative depth estimation on the 4th test image are shown in Fig. \ref{real4}(b-f), Note that the three different planes are clearly distinguishable in the final iteration. Again, it can be seen that our approach recovers scene depth and texture faithfully, while the results of existing methods contain visible artifacts in inclined regions.

\section{Conclusions}
We formulated the underlying relationship between the surface normal of a planar scene and the induced space-variant nature of blur due to camera motion. By utilizing the correspondences among the extreme points of the PSFs, we proposed a new approach to solve for the surface normal of a planar scene. The method leads to robust normal estimation even on real images which can be conveniently plugged into existing image formation model for restoration of motion-blurred 3D scenes. Finally, we proposed a first-of-its-kind scheme to estimate orientation of multiple planes from a single motion blurred image and utilized it to deblur the image. Our proposed approach achieves state-of-the-art results for the task of single image 3D scene motion deblurring.

\noindent Refined and complete version of this work appeared in the Journal of Machine Vision and Applications 2021.

\bibliographystyle{unsrt}  
\bibliography{refs_5}

\begin{thebibliography}{10}

\bibitem{lee2011real}
Sang~Hwa Lee and Siddharth Sharma.
\newblock Real-time disparity estimation algorithm for stereo camera systems.
\newblock {\em IEEE transactions on Consumer electronics}, 57(3), 2011.

\bibitem{shahraray1988robust}
Behzad Shahraray and Michael~K Brown.
\newblock Robust depth estimation from optical flow.
\newblock In {\em Computer Vision., Second International Conference on}, pages
  641--650. IEEE, 1988.

\bibitem{super1995planar}
Boaz~J Super and Alan~C Bovik.
\newblock Planar surface orientation from texture spatial frequencies.
\newblock {\em Pattern Recognition}, 28(5):729--743, 1995.

\bibitem{brown1990surface}
Lisa~Gottesfeld Brown and Haim Shvaytser.
\newblock Surface orientation from projective foreshortening of isotropic
  texture autocorrelation.
\newblock {\em IEEE Transactions on Pattern Analysis and Machine Intelligence},
  12(6):584--588, 1990.

\bibitem{zhang1999shape}
Ruo Zhang, Ping-Sing Tsai, James~Edwin Cryer, and Mubarak Shah.
\newblock Shape-from-shading: a survey.
\newblock {\em IEEE transactions on pattern analysis and machine intelligence},
  21(8):690--706, 1999.

\bibitem{chaudhuri2012depth}
Subhasis Chaudhuri and Ambasamudram~N Rajagopalan.
\newblock {\em Depth from defocus: a real aperture imaging approach}.
\newblock Springer Science \& Business Media, 2012.

\bibitem{chandramouli2010inferring}
Paramanand Chandramouli and A~Rajagopalan.
\newblock Inferring image transformation and structure from motion-blurred
  images.
\newblock In {\em BMVC}, pages 73--1, 2010.

\bibitem{lin2006depth}
Huei-Yung Lin and Chia-Hong Chang.
\newblock Depth recovery from motion blurred images.
\newblock In {\em Pattern Recognition, 2006. ICPR 2006. 18th International
  Conference on}, volume~1, pages 135--138. IEEE, 2006.

\bibitem{zheng2011structure}
Yali Zheng, Shohei Nobuhara, and Yaser Sheikh.
\newblock Structure from motion blur in low light.
\newblock In {\em CVPR}, pages 2569--2576. IEEE, 2011.

\bibitem{clark2001estimating}
Paul Clark and Majid Mirmehdi.
\newblock Estimating the orientation and recovery of text planes in a single
  image.
\newblock In {\em BMVC}, pages 1--10, 2001.

\bibitem{farid2007estimating}
Hany Farid and Jana Kosecka.
\newblock Estimating planar surface orientation using bispectral analysis.
\newblock {\em IEEE Transactions on image processing}, 16(8):2154--2160, 2007.

\bibitem{greiner2010estimation}
Thomas Greiner, Shivani~G Rao, and Sukhendu Das.
\newblock Estimation of orientation of a textured planar surface using
  projective equations and separable analysis with m-channel wavelet
  decomposition.
\newblock {\em Pattern Recognition}, 43(1):230--243, 2010.

\bibitem{mccloskey2009planar}
Scott McCloskey and Michael Langer.
\newblock Planar orientation from blur gradients in a single image.
\newblock In {\em CVPR}, pages 2318--2325. IEEE, 2009.

\bibitem{rao2014inferring}
M~Purnachandra Rao, AN~Rajagopalan, and Guna Seetharaman.
\newblock Inferring plane orientation from a single motion blurred image.
\newblock In {\em ICPR}, pages 2089--2094. IEEE, 2014.

\bibitem{vasu2015tapping}
Subeesh Vasu, AN~Rajagopalan, and Gunasekaran Seetharaman.
\newblock Tapping motion blur for robust normal estimation of planar scenes.
\newblock In {\em ICIP}, pages 2761--2765. IEEE, 2015.

\bibitem{saxena2009make3d}
Ashutosh Saxena, Min Sun, and Andrew~Y Ng.
\newblock Make3d: Learning 3d scene structure from a single still image.
\newblock {\em IEEE transactions on pattern analysis and machine intelligence},
  31(5):824--840, 2009.

\bibitem{haines2012detecting}
Osian Haines and Andrew Calway.
\newblock Detecting planes and estimating their orientation from a single
  image.
\newblock In {\em BMVC}, pages 1--11, 2012.

\bibitem{eigen2014depth}
David Eigen, Christian Puhrsch, and Rob Fergus.
\newblock Depth map prediction from a single image using a multi-scale deep
  network.
\newblock In {\em NIPS}, pages 2366--2374, 2014.

\bibitem{laina2016deeper}
Iro Laina, Christian Rupprecht, Vasileios Belagiannis, Federico Tombari, and
  Nassir Navab.
\newblock Deeper depth prediction with fully convolutional residual networks.
\newblock In {\em 3D Vision (3DV), 2016 Fourth International Conference on},
  pages 239--248. IEEE, 2016.

\bibitem{li2018megadepth}
Zhengqi Li and Noah Snavely.
\newblock Megadepth: Learning single-view depth prediction from internet
  photos.
\newblock In {\em Proceedings of the IEEE Conference on Computer Vision and
  Pattern Recognition}, pages 2041--2050, 2018.

\bibitem{rao2014harnessing}
Makkena~Purnachandra Rao, AN~Rajagopalan, and Guna Seetharaman.
\newblock Harnessing motion blur to unveil splicing.
\newblock {\em IEEE transactions on information forensics and security},
  9(4):583--595, 2014.

\bibitem{mohan2019unconstrained}
MR~Mohan, Sharath Girish, and AN~Rajagopalan.
\newblock Unconstrained motion deblurring for dual-lens cameras.
\newblock In {\em Proceedings of the IEEE/CVF International Conference on
  Computer Vision}, pages 7870--7879, 2019.

\bibitem{nimisha2018generating}
TM~Nimisha, AN~Rajagopalan, and Rangarajan Aravind.
\newblock Generating high quality pan-shots from motion blurred videos.
\newblock {\em Computer Vision and Image Understanding}, 171:20--33, 2018.

\bibitem{vasu2017local}
Subeesh Vasu and AN~Rajagopalan.
\newblock From local to global: Edge profiles to camera motion in blurred
  images.
\newblock In {\em Proceedings of the IEEE Conference on Computer Vision and
  Pattern Recognition}, pages 4447--4456, 2017.

\bibitem{paramanand2014shape}
Chandramouli Paramanand and AN~Rajagopalan.
\newblock Shape from sharp and motion-blurred image pair.
\newblock {\em International journal of computer vision}, 107(3):272--292,
  2014.

\bibitem{paramanand2011depth}
Chandramouli Paramanand and Ambasamudram~N Rajagopalan.
\newblock Depth from motion and optical blur with an unscented kalman filter.
\newblock {\em IEEE Transactions on Image Processing}, 21(5):2798--2811, 2011.

\bibitem{vijay2013non}
Channarayapatna~Shivaram Vijay, Chandramouli Paramanand, Ambasamudram~Narayanan
  Rajagopalan, and Rama Chellappa.
\newblock Non-uniform deblurring in hdr image reconstruction.
\newblock {\em IEEE transactions on image processing}, 22(10):3739--3750, 2013.

\bibitem{nimisha2018unsupervised}
Thekke~Madam Nimisha, Kumar Sunil, and AN~Rajagopalan.
\newblock Unsupervised class-specific deblurring.
\newblock In {\em Proceedings of the European Conference on Computer Vision
  (ECCV)}, pages 353--369, 2018.

\bibitem{purohit2020region}
Kuldeep Purohit and AN~Rajagopalan.
\newblock Region-adaptive dense network for efficient motion deblurring.
\newblock In {\em Proceedings of the AAAI Conference on Artificial
  Intelligence}, volume~34, pages 11882--11889, 2020.

\bibitem{vasu2018non}
Subeesh Vasu, Venkatesh~Reddy Maligireddy, and AN~Rajagopalan.
\newblock Non-blind deblurring: Handling kernel uncertainty with cnns.
\newblock In {\em Proceedings of the IEEE Conference on Computer Vision and
  Pattern Recognition}, pages 3272--3281, 2018.

\bibitem{purohit2019bringing}
Kuldeep Purohit, Anshul Shah, and AN~Rajagopalan.
\newblock Bringing alive blurred moments.
\newblock In {\em Proceedings of the IEEE/CVF Conference on Computer Vision and
  Pattern Recognition}, pages 6830--6839, 2019.

\bibitem{rajagopalan2005background}
AN~Rajagopalan, Rama Chellappa, and Nathan~T Koterba.
\newblock Background learning for robust face recognition with pca in the
  presence of clutter.
\newblock {\em IEEE Transactions on Image Processing}, 14(6):832--843, 2005.

\bibitem{fergus2006removing}
Rob Fergus, Barun Singh, Aaron Hertzmann, Sam~T Roweis, and William~T Freeman.
\newblock Removing camera shake from a single photograph.
\newblock In {\em ACM transactions on graphics (TOG)}, volume~25, pages
  787--794. ACM, 2006.

\bibitem{cho2009fast}
Sunghyun Cho and Seungyong Lee.
\newblock Fast motion deblurring.
\newblock In {\em ACM Transactions on Graphics (TOG)}, volume~28, page 145.
  ACM, 2009.

\bibitem{xu2010two}
Li~Xu and Jiaya Jia.
\newblock Two-phase kernel estimation for robust motion deblurring.
\newblock {\em ECCV}, pages 157--170, 2010.

\bibitem{sun2013edge}
Libin Sun, Sunghyun Cho, Jue Wang, and James Hays.
\newblock Edge-based blur kernel estimation using patch priors.
\newblock In {\em ICCP}, pages 1--8. IEEE, 2013.

\bibitem{michaeli2014blind}
Tomer Michaeli and Michal Irani.
\newblock Blind deblurring using internal patch recurrence.
\newblock In {\em ECCV}, pages 783--798. Springer, 2014.

\bibitem{gupta2010single}
Ankit Gupta, Neel Joshi, C~Lawrence~Zitnick, Michael Cohen, and Brian Curless.
\newblock Single image deblurring using motion density functions.
\newblock {\em ECCV}, pages 171--184, 2010.

\bibitem{hirsch2011fast}
Michael Hirsch, Christian~J Schuler, Stefan Harmeling, and Bernhard
  Sch{\"o}lkopf.
\newblock Fast removal of non-uniform camera shake.
\newblock In {\em ICCV}, pages 463--470. IEEE, 2011.

\bibitem{whyte2012non}
Oliver Whyte, Josef Sivic, Andrew Zisserman, and Jean Ponce.
\newblock Non-uniform deblurring for shaken images.
\newblock {\em International journal of computer vision}, 98(2):168--186, 2012.

\bibitem{Vasu_2017_CVPR}
Subeesh Vasu and A.~N. Rajagopalan.
\newblock From local to global: Edge profiles to camera motion in blurred
  images.
\newblock In {\em CVPR}, July 2017.

\bibitem{Yan_2017_CVPR}
Yanyang Yan, Wenqi Ren, Yuanfang Guo, Rui Wang, and Xiaochun Cao.
\newblock Image deblurring via extreme channels prior.
\newblock In {\em The IEEE Conference on Computer Vision and Pattern
  Recognition (CVPR)}, July 2017.

\bibitem{hu2014joint}
Zhe Hu, Li~Xu, and Ming-Hsuan Yang.
\newblock Joint depth estimation and camera shake removal from single blurry
  image.
\newblock In {\em CVPR}, pages 2893--2900, 2014.

\bibitem{seemakurthy2016deskewing}
Karthik Seemakurthy, Subeesh Vasu, and Rajagopalan Ambasamudram.
\newblock Deskewing by space-variant deblurring.
\newblock In {\em BMVC}, 2016.

\bibitem{sorel2008space}
Michal Sorel and Jan Flusser.
\newblock Space-variant restoration of images degraded by camera motion blur.
\newblock {\em IEEE Transactions on Image Processing}, 17(2):105--116, 2008.

\bibitem{xu2012depth}
Li~Xu and Jiaya Jia.
\newblock Depth-aware motion deblurring.
\newblock In {\em ICCP}, pages 1--8. IEEE, 2012.

\bibitem{paramanand2013non}
Chandramouli Paramanand and Ambasamudram~N Rajagopalan.
\newblock Non-uniform motion deblurring for bilayer scenes.
\newblock In {\em CVPR}, pages 1115--1122, 2013.

\bibitem{sun2015learning}
Jian Sun, Wenfei Cao, Zongben Xu, Jean Ponce, et~al.
\newblock Learning a convolutional neural network for non-uniform motion blur
  removal.
\newblock In {\em CVPR}, pages 769--777, 2015.

\bibitem{gong2017motion}
Dong Gong, Jie Yang, Lingqiao Liu, Yanning Zhang, Ian Reid, Chunhua Shen, AVD
  Hengel, and Qinfeng Shi.
\newblock From motion blur to motion flow: a deep learning solution for
  removing heterogeneous motion blur.
\newblock In {\em CVPR}, 2017.

\bibitem{nimisha2017blur}
TM~Nimisha, Akash~Kumar Singh, and AN~Rajagopalan.
\newblock Blur-invariant deep learning for blind-deblurring.
\newblock In {\em ICCV}, pages 4752--4760, 2017.

\bibitem{nah2017deep}
Seungjun Nah, Tae~Hyun Kim, and Kyoung~Mu Lee.
\newblock Deep multi-scale convolutional neural network for dynamic scene
  deblurring.
\newblock In {\em CVPR}, volume 2017, 2017.

\bibitem{tao2018scale}
Xin Tao, Hongyun Gao, Xiaoyong Shen, Jue Wang, and Jiaya Jia.
\newblock Scale-recurrent network for deep image deblurring.
\newblock {\em arXiv preprint arXiv:1802.01770}, 2018.

\bibitem{kohler2012recording}
Rolf K{\"o}hler, Michael Hirsch, Betty Mohler, Bernhard Sch{\"o}lkopf, and
  Stefan Harmeling.
\newblock Recording and playback of camera shake: Benchmarking blind
  deconvolution with a real-world database.
\newblock {\em ECCV}, pages 27--40, 2012.

\bibitem{boyd2011distributed}
Stephen Boyd, Neal Parikh, Eric Chu, Borja Peleato, and Jonathan Eckstein.
\newblock Distributed optimization and statistical learning via the alternating
  direction method of multipliers.
\newblock {\em Foundations and Trends in Machine Learning}, 3(1):1--122, 2011.

\bibitem{wang2008new}
Yilun Wang, Junfeng Yang, Wotao Yin, and Yin Zhang.
\newblock A new alternating minimization algorithm for total variation image
  reconstruction.
\newblock {\em SIAM Journal on Imaging Sciences}, 1(3):248--272, 2008.

\bibitem{boykov2001fast}
Yuri Boykov, Olga Veksler, and Ramin Zabih.
\newblock Fast approximate energy minimization via graph cuts.
\newblock {\em IEEE Transactions on pattern analysis and machine intelligence},
  23(11):1222--1239, 2001.

\bibitem{fischler1981random}
Martin~A Fischler and Robert~C Bolles.
\newblock Random sample consensus: a paradigm for model fitting with
  applications to image analysis and automated cartography.
\newblock {\em Communications of the ACM}, 24(6):381--395, 1981.

\bibitem{xu2011image}
Li~Xu, Cewu Lu, Yi~Xu, and Jiaya Jia.
\newblock Image smoothing via l 0 gradient minimization.
\newblock In {\em ACM Transactions on Graphics (TOG)}, volume~30, page 174.
  ACM, 2011.

\bibitem{kupyn2018deblurgan}
Orest Kupyn, Volodymyr Budzan, Mykola Mykhailych, Dmytro Mishkin, and
  Ji{\v{r}}{\'\i} Matas.
\newblock Deblurgan: Blind motion deblurring using conditional adversarial
  networks.
\newblock In {\em Proceedings of the IEEE Conference on Computer Vision and
  Pattern Recognition}, pages 8183--8192, 2018.

\bibitem{Sun:2012:SR_sun_hays}
Libin Sun and James Hays.
\newblock Super-resolution from internet-scale scene matching.
\newblock In {\em Proceedings of the {IEEE} Conf. on International Conference
  on Computational Photography ({ICCP})}, 2012.

\end{thebibliography}

\end{document}